\documentclass[runningheads]{llncs}

\usepackage[mobile]{eccv}

\usepackage{eccvabbrv}

\usepackage{graphicx}
\usepackage{booktabs}
\usepackage{colortbl}
\usepackage{algorithm}
\usepackage{algpseudocode}
\usepackage{amsmath}
\usepackage{wrapfig}
\usepackage{tabularx}
\usepackage{multirow}
\usepackage{listings}
\usepackage{enumitem}

\usepackage[accsupp]{axessibility}  

\usepackage{hyperref}

\usepackage{orcidlink}

\definecolor{codeblue}{RGB}{37, 105, 188}
\definecolor{codegray}{RGB}{128, 128, 128}
\definecolor{codepurple}{RGB}{145, 75, 180}

\lstnewenvironment{python}[1][]{
    \lstset{
        language=Python,
        basicstyle=\ttfamily\small,
        keywordstyle=\color{codeblue}\bfseries,
        commentstyle=\color{codegray},
        showstringspaces=false,
        breaklines=true,
        frame=none,
        mathescape=true, 
        #1
    }
}{}

\begin{document}


\title{Don’t Settle at the Mode! \\ Mitigating Diversity Collapse in Pretrained Flow Models via Feature Self-Guidance}

\titlerunning{Don't Settle at the Mode!}

\author{
Pradhaan S Bhat\inst{1}\thanks{Equal Contribution} \and
Rishubh Parihar\inst{1}\protect\footnotemark[1] \and
Abhijnya Bhat\inst{2} \and \\
R. Venkatesh Babu\inst{1}
}

\authorrunning{P.Bhat et al.}

\institute{Indian Institute of Science \and
Stanford University}


\maketitle
\begin{center}
\vspace{-2mm}
\textbf{Project Page:} \texorpdfstring{\url{https://dont-settle-at-the-mode.github.io/}}{https://dont-settle-at-the-mode.github.io/}
\end{center}

\begin{figure*}
    \vspace{-8mm}
  \centering
   \includegraphics[width=1.0\linewidth]{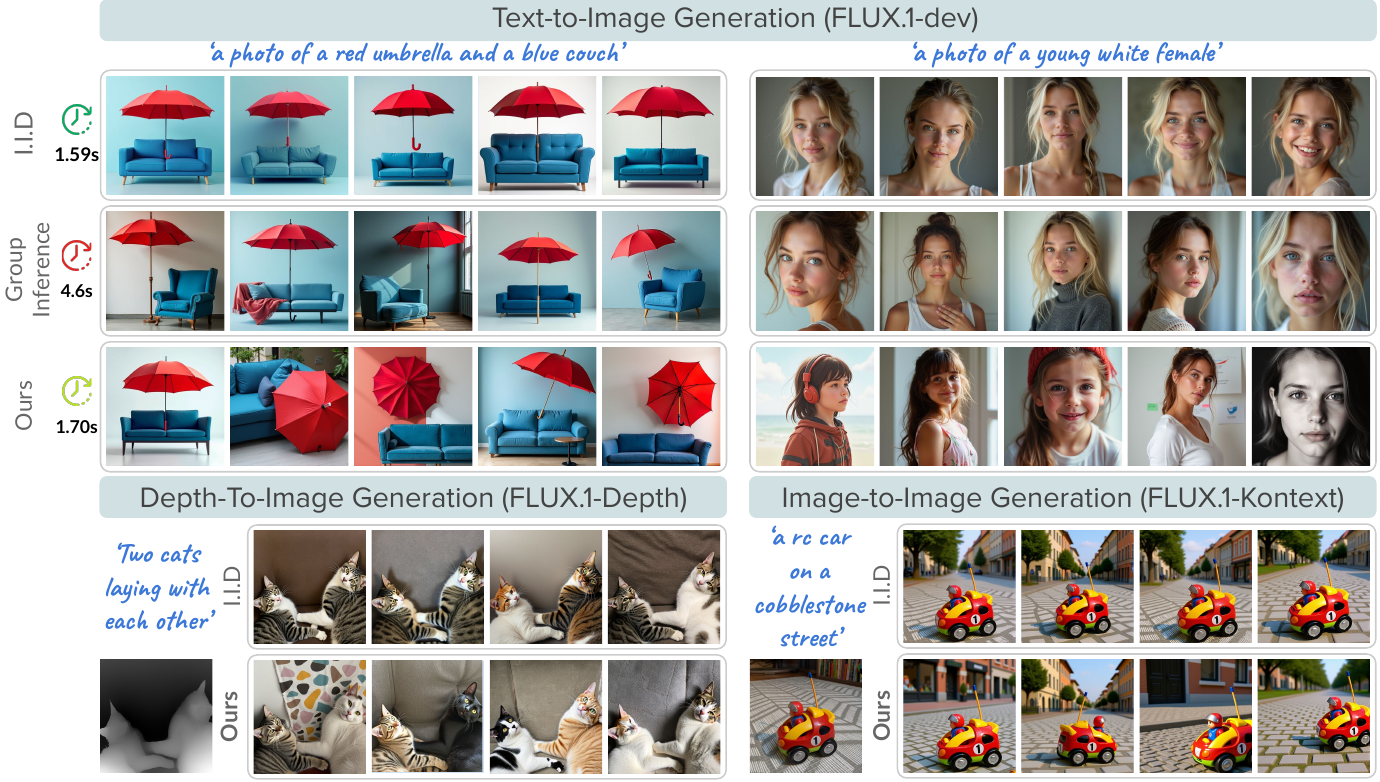}
   \caption{We propose an efficient inference-time approach to enhance diversity in conditional flow models. Our plug-and-play module integrates seamlessly into various conditional flow models and consistently enhances diversity for text-to-image, depth-to-image and image-to-image models. Compared to recent Group Inference~\cite{group-inference}, which relies on reward-based sample rejection, our method achieves competitive diversity without external reward models while requiring only a fraction of their inference time.} 
   \vspace{-8mm}
   \label{fig:teaser}
\end{figure*}

\begin{abstract}
State-of-the-art flow models generate stunning images from text or image prompts. However, they suffer from diversity collapse when generating multiple samples under the same conditioning. Existing methods address this issue via either latent guidance, which has limited effectiveness, or sample selection, which relies on external reward models that incur significant inference-time overhead. In this work, we introduce an efficient, training-free self-guidance mechanism to mitigate diversity collapse without requiring additional reward models. Specifically, we disperse the internal features of the flow model during batch generation with \textbf{feature self-guidance}. Further, to keep the features close to the manifold, we introduce a \textbf{manifold regularization} step that projects these dispersed features back onto the data manifold, ensuring diverse generation without sacrificing alignment with the input conditions. Our method integrates seamlessly as a plug-and-play module into pretrained flow models, adding only a marginal inference cost. Experiments demonstrate significant improvements in diversity while preserving fidelity across several conditional flow models, including multi-step and few-step text-to-image, depth-to-image, and reference image generation. 

\end{abstract}
\section{Introduction}
\label{sec:intro}

While state-of-the-art flow-based generative models~\cite{sd3, flux, flux-2, qwen-image} achieve exceptional generation quality, they often exhibit a lack of sample diversity when prompted with the same conditioning~\cite{group-inference, nevertoolate, shielded-diff}. As illustrated in Fig.~\ref{fig:teaser}, widely used FLUX.1-dev~\cite{flux} produces nearly redundant outputs for the prompt \textit{`a portrait photo of a woman'}, capturing only a narrow subset of the potential distribution. Diverse generation is critical in practical applications, where a wide array of outputs is necessary to facilitate user choice and provide inspiration for iterative prompt engineering.

\setlength{\intextsep}{2pt}
\setlength{\columnsep}{8pt}
\begin{wrapfigure}{r}{0.5\textwidth}
    \vspace{-6mm}
  \centering
   \includegraphics[width=0.75\linewidth]{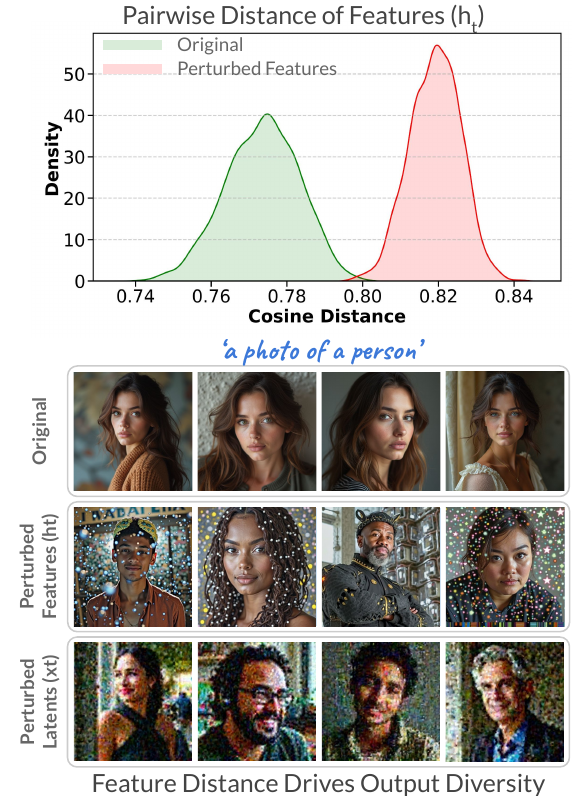}
    \vspace{-2mm}
   \caption{\textbf{Feature Distance vs. Output Diversity.} We examine the impact of perturbing FLUX model's internal DiT features $h_t$ with Gaussian noise. As shown in the histograms, increasing the pairwise distance between internal features increases sample diversity. In contrast, perturbing latents $x_t$ with noise causes image corruption. This identifies feature dispersion as a mechanism for recovering flow model diversity.}
   \label{fig:motivation}
\end{wrapfigure}

Existing methods to mitigate diversity collapse in pretrained flow models during inference generally fall into two categories. The first focuses on latent guidance~\cite{shielded-diff, particle-guidance, interval-guidance, CNO}, which guides latents to generate varied samples. While computationally efficient, these approaches often yield marginal improvements in sample variety. The second category leverages external reward models to either select the most diverse samples from a candidate set~\cite{group-inference} or optimize the initial noise for enhanced variety~\cite{nevertoolate}. While these reward-driven strategies significantly enhance diversity, they introduce substantial computational overhead due to the repeated inference of reward models (Fig.~\ref{fig:teaser}) over decoded latents in the pixel space. In this work, we introduce an efficient and effective method for recovering diversity in pretrained flow models without external models.

Our key insight is that \textit{diversity collapse can be mitigating by addressing the collapse of internal features of the flow models}. To validate this hypothesis, we conduct a diagnostic experiment where we intentionally expand the internal representation space of the pretrained FLUX.1-dev model. Specifically, we introduce a stochastic perturbation to the intermediate Multimodal Diffusion Transformer~\cite{sd3} (MMDiT) block features $h_t$ by injecting Gaussian noise, defined as $\hat{h}_t = h_t + \epsilon$, where $\epsilon \sim \mathcal{N}(0,\sigma^2 I)$. This controlled setup allows us to observe how shifts in the feature space affect the resulting sample diversity. As illustrated in the pairwise distance histograms and qualitative results in ~\cref{fig:motivation}, increasing the feature-space variance directly leads to a significant rise in the diversity of generated samples. This reveals a clear correspondence: \textit{the spatial dispersion of internal MMDiT features serves as a direct proxy for the diversity of the final image outputs.} Interestingly, the simple alternate of latent perturbations ($x_t$) degrades image quality, causing the loss of high-frequency and texture details and resulting in severe image corruption, whereas perturbing internal features largely preserves image structure and details, producing only mild overlaying artifacts. This highlights the greater robustness of the feature space as a target for diversity enhancement. However, unconstrained perturbation often drives features into low-density regions, manifesting as visual artifacts. To address this, we propose a principled dispersion of internal features expanding diversity while anchoring them to the feature manifold to preserve image fidelity and faithfulness.

To maximize the variance of internal MMDiT features without compromising semantic integrity, we introduce an efficient \textbf{feature self-guidance} mechanism. For a given batch of latents, we extract intermediate features $h_t$ and apply a dispersion guidance that pushes the features apart. However, unconditional dispersion risks pushing features beyond the semantic boundary, risking the alignment with the input conditioning. To mitigate this, we propose a \textbf{manifold regularization} where the dispersed features are reprocessed through the same MMDiT block to project them back toward the valid conditional feature manifold. The projected features are then used as a regularizer for the raw dispersed features. Through rigorous analysis, we demonstrate that applying this guidance to a single MMDiT block over a selected timestep window is sufficient to maximize diversity while maintaining a minimal computational footprint.

Our training-free approach is designed for seamless integration into any pretrained flow model as a plug-and-play module. To demonstrate its broad applicability, we conduct extensive evaluations across a diverse set of conditional flow models: FLUX.1 for text-to-image generation, FLUX.1-Depth for depth-conditioned synthesis, FLUX-Kontext for reference image generation, and the step-distilled FLUX.2-Klein. Across all benchmarks, our method significantly outperforms existing diversity-enhancement baselines while preserving the native high-speed inference characteristic of the underlying flow models. 

\vspace{2mm}
In summary, our key contributions are:

\begin{itemize}
\item The key discovery of a direct correlation between internal feature collapse in pretrained flow models and output diversity collapse.

\item Feature Self-Guidance that effectively disperse of internal features of the flow model to enhance diversity while utilizing manifold regularization to anchor samples to the conditioning signal and preserve image fidelity.

\item Extensive empirical validation across diverse tasks—including state-of-the-art text-to-image and conditional generation, demonstrating that our method consistently recovers diversity without sacrificing image quality.
\end{itemize}

\section{Related Works}

\vspace{2mm}
\noindent\textbf{Guidance and Diversity:} 
Standard Classifier-Free Guidance~\cite{CFG} can be used to control the diversity in pretrained diffusion models~\cite{ddpm,ddim,score-sde}. However, it has a poor faithfulness and diversity tradeoff, where highly diverse samples don't follow the input conditioning. Several guidance-based inference strategies have emerged
to steer the model towards a better fidelity-diversity trade-off~\cite{ERG,Sparke}. Balancing-Act~\cite{balancing-act} guides the internal h-space features of UNet based diffusion models to improve attribute diversity. Interval Guidance~\cite{interval-guidance} balances these objectives by restricting the guidance to specific inference windows. Other approaches focus on the sampling trajectory itself: Particle Guidance~\cite{particle-guidance} learns a joint-particle time-evolving potential to maximize diversity, while Shielded Diffusion~\cite{shielded-diff} promotes diversity by sparsely repelling noisy latents at some inference steps. Alternatively, CNO~\cite{CNO} bypasses per-step guidance entirely by optimizing the initial latent space. Unlike methods operating in the latent space, our approach operates directly in the internal feature space of MMDiT~\cite{sd3} blocks. By shifting the intervention to the internal feature space, we provide an effective solution by mitigating collapse at its source. 

\vspace{2mm}
\noindent\textbf{Prompt Optimization for Diversity:} Improving generative control by optimizing the input prompt embedding without training the flow/diffusion model is a well-established direction~\cite{galimage,alaluf2023neural,gal2023encoder}. In similar lines, CADS~\cite{cads} enhances diversity by perturbing the conditioning signal; however, this often compromises prompt adherence and leads to semantic drift. Other methods directly optimize~\cite{minority-prompt,learning-to-sample-diverse-prompts,DreamDistribution,Dual-process} for diversity or employ multimodal large language models to iteratively refine the prompt based on visual feedback~\cite{ReflectionFlow,TIR}. Because these methods necessitate test-time optimization, their practical application is constrained; moreover, optimization within the text space lacks the expressivity required to fully capture the complexity of diverse data distributions.

\vspace{2mm}
\noindent\textbf{Representation Regularization for Generation:} Recent works have leveraged self-supervised learning to introduce intermediate representation regularization during the training of Diffusion Transformers~\cite{repa,rae,scaling-rae,irepa,diffuse-disperse}. These methods, which primarily aim to enhance training efficiency, can be broadly categorized into two strategies: (1) aligning internal representations of the flow models with the pre-trained self-supervised models~\cite{repa, irepa, rae, scaling-rae}, and (2) incorporating self-supervised learning as an auxiliary regularization task during flow model training~\cite{diffuse-disperse, internal-guidance}. Our method takes inspiration from the latter set of methods; however, we modify the internal representations during inference via feature guidance without the need for training.

\vspace{2mm}
\noindent\textbf{Inference-time Scaling}. Inference-time scaling~\cite{llm-scaling,CoT} in diffusion and flow models improves sample quality by allocating more computational resources during inference, like having to reward the model or performing optimization. There are two primary directions in inference-time scaling: iterative optimization and discrete selection. Optimization-based methods refine the initial Gaussian noise for point-wise quality via quality rewards~\cite{reno,dno,noise-hypernetworks} along with set-level objectives to explicitly manage diversity~\cite{nevertoolate} or integrate Determinantal Point Processes (DPP) into policy optimization loops~\cite{dpp-grpo-video}. On the other hand, selection-based approaches leverage reward models to identify optimal trajectories through verifier feedback. These methods scale performance either by filtering a large pool of sample candidates to identify `Best-of-N' samples~\cite{Saining-inference} or by sampling from a reward-tilted distribution~\cite{Minhyuk-inference}. This setting extends to set-wise diverse selection strategies via the use of Quadratic Integer Programming at intermediate denoising steps to balance quality and diversity within a generated group of samples~\cite{group-inference}. However, a significant bottleneck shared by both types of approaches is their reliance on reward models, significantly improving the number of function evaluations during inference. In contrast, we demonstrate that high-quality, diverse samples can be produced efficiently by \textit{dispersing} flow features on a constrained manifold, bypassing the need for additional reward supervision.
\section{Method}

\begin{figure}[t]
    \centering
    \includegraphics[width=\linewidth]{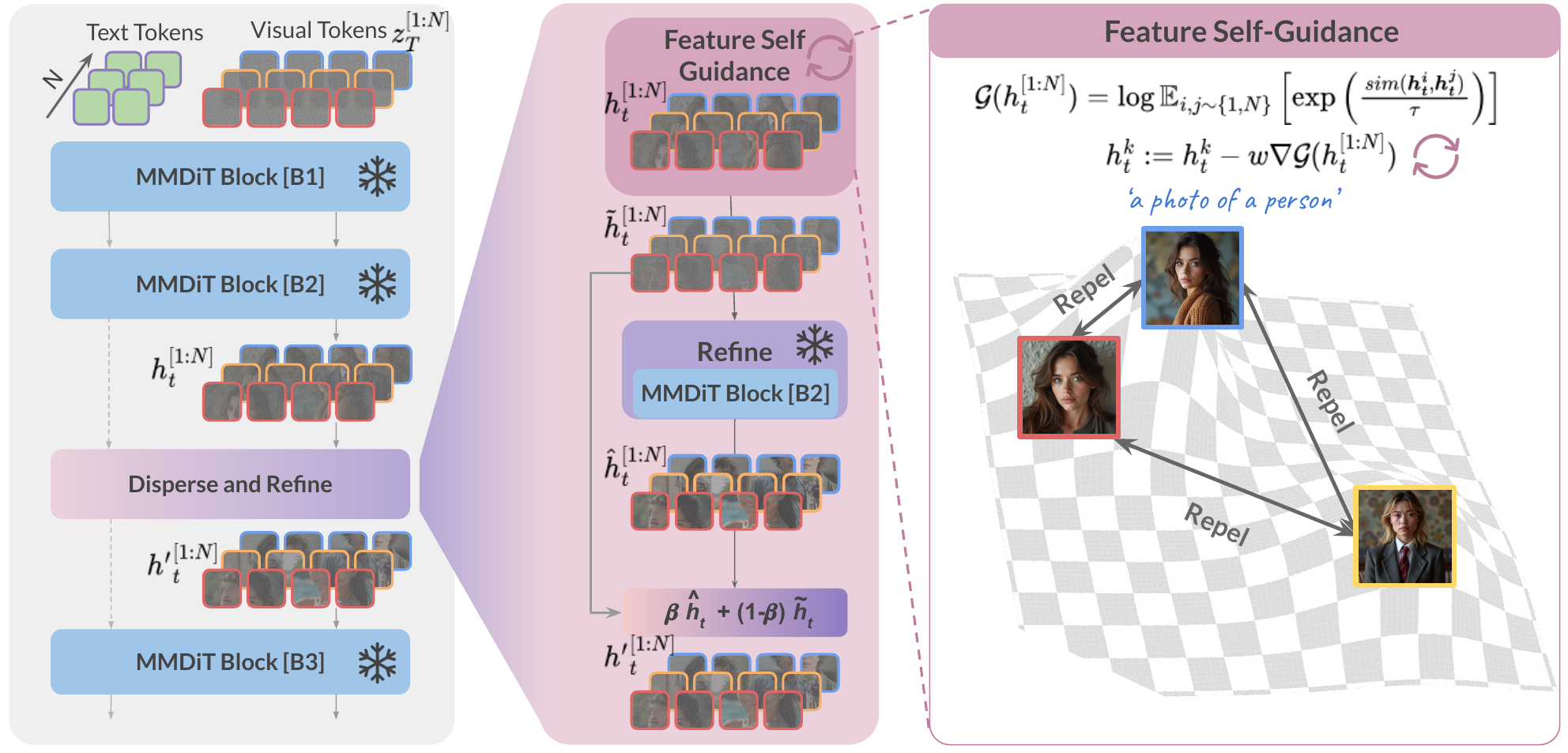}
    \caption{\textbf{Overview of Feature Self-Guidance.} Our method integrates a "Disperse-and-Refine" module into the flow model backbone. \textbf{(a)} Dispersion Stage: Internal features are pushed apart using iterative self-guidance to expand sample variety. \textbf{(b)} Refinement Stage: These features are re-processed through the same MMDiT block to project them back onto the valid image manifold. Finally, we linearly interpolate the dispersed and projected features to ensure a high-fidelity and, diverse output.} 
    \vspace{-2mm}
    \label{fig:2_method}
\end{figure}

\subsection{Preliminaries}
\noindent 
\textbf{Flow Models.} Flow models~\cite{flow-matching,rectified-flow} learn a continuous transformation from a Gaussian prior $p_1 \sim \mathcal{N}(0,I)$ to a target data distribution $p_0$ by modeling an Ordinary Differential Equation (ODE) trajectory. For training, these models define an interpolant $x_t$ between a data point $x_0 \sim p_0(x)$ and a noise sample $x_T \sim p_1$ as $x_t$ = $(1-t) x_0 + t x_T$ for $t \in \mathcal{U}[0,1]$. A neural network $v_\theta$ is trained to predict the conditional velocity field $v_\theta (x_t,t,c)$ to match the true velocity $(x_T - x_0)$. The training objective, known as the Conditional Flow Matching (CFM) loss, is defined as:

\begin{equation}
\mathcal{L}_{CFM}(\theta) = \mathbb{E}_{t \sim \mathcal{U}[0, 1], x_0 \sim p_0, x_T \sim \mathcal{N}(0, \mathbf{I})} \left[ \left\| v_\theta(x_t, t, c) - (x_T-x_0) \right\|^2 \right]
\end{equation}

\noindent 
where $c$ represents the conditioning signal such as text prompts.

\vspace{1mm}
\noindent 
\textbf{Guidance.} 
Analogous to classifier guidance in diffusion models~\cite{classifier-guidance}, flow models can be adapted for inference-time conditioning by steering the predicted score $\nabla\log p_t(x_t)$ via a pretrained classifier $p(c|x_t)$ as follows:
\begin{equation}
\nabla\log p_t(x_t|c) = \nabla\log p_t(x_t) + \lambda\nabla_{x_{t}}p(c|x_t)
\end{equation}
where $c$ is the class label. In principle, this guidance can be applied using any differentiable energy function $\mathcal{G}(x_t,c,t)$, such as CLIP~\cite{clip} for enhancing prompt-adherence, an object detector for layout-conditioned generation~\cite{bansal2023universal}, or sketches for sketch-to-image synthesis~\cite{voynov2023sketch}. Alternatively, the energy function can be computed directly from the model's internal features via self-guidance~\cite{epstein2023diffusion}. Because self-guidance bypasses external model evaluations, it only adds a marginal computational cost to the inference process.

\subsection{Overview}

We build on FLUX~\cite{flux}, a DiT architecture for text-to-image (T2I) generation which comprises of a series of multimodal DiT (MMDiT) blocks $B_1 \cdots B_{57}$ ~\cite{sd3}. These blocks jointly process text and image tokens through self-attention and feed-forward layers facilitating rich information exchange between text and image tokens. Leveraging the rich information within the MMDiT blocks, we introduce a strategy to \textbf{disperse} the image features and mitigate mode collapse. To ensure these \textbf{dispersed} features remain on the data manifold, we incorporate a \textbf{refinement} step that regularizes the features, preventing them from drifting into low-probability regions. This dual \textbf{disperse-and-refine} mechanism enhances generative diversity without sacrificing fidelity. Detailed ablations and design choices follow in the subsequent sections. We present the overall algorithm in Algorithm \ref{alg:feature-guidance}.

\subsection{Feature Self-Guidance for diversity}

We build our method on the insight from ~\cref{fig:motivation}, i.e., the diversity collapse in the generated samples is due to the collapse of the internal MMDiT features $h_t$. We design a principled method that effectively disperses the internal MMDiT features while preserving the image fidelity. We formulate a feature self-guidance method that iteratively disperses the batch sample features $h_t^{[1:N]}$ through guidance. We use pair-wise feature distance to formulate our self-guidance energy function $\mathcal{G}$ defined as: 
\begin{equation}
\label{eqn:1}
\mathcal{G}(h_t^{[1:N]}) = \log \mathbb{E}_{i,j \sim \{1,N\}} \left[ \exp\left( \frac{sim(\boldsymbol{h_{t}^i}, \boldsymbol{h_{t}^j})}{\tau} \right) \right]
\end{equation}

\noindent 
where $N$ is the number of samples in a batch and $sim(\mathbf{h}_{t}^{i}, \mathbf{h}_{t}^{j})$ is the cosine similarity between the features. However, unlike existing guidance-based approaches that update the latents $x_t$ with gradients of energy function $\nabla \mathcal{G}(h_t)$, we directly update the internal features $h_t$ as below, enforcing feature diversity more explicitly by pushing the features apart: 
\begin{equation}
\label{eqn:2}
    h_t^k := h_t^k - w \nabla \mathcal{G}(h_t^{[1:N]}) 
\end{equation} 

\noindent 
Notably, updating $h_t$ is much faster than updating $x_t$ with guidance, as it only modifies intermediate features without requiring backpropagation through previous transformer blocks. This efficient self-guidance enables diverse generations; however, it may reduce faithfulness to onditioning, as iterative dispersion may push features outside the feature manifold. To address this, we propose a manifold regularization step that keeps updated features in distribution (Sec.~\ref{subsec:manifold}).

\vspace{2mm}
\noindent 
\textbf{Which block is most effective for feature self-guidance?}
\vspace{2mm}

\noindent 
Inspired by StableFlow~\cite{StableFlow}, which shows that early MMDiT blocks capture most semantic information, we evaluate feature guidance across the initial blocks (Fig.~\ref{fig:8_block}). Block $B_2$ performs best, while adding it to later blocks yields only marginal diversity gains. Furthermore, since image structure is primarily formed in the early denoising steps, we apply guidance only for $t \in [1.0, 0.8]$. Additional ablations on block and timestep selection are provided in the appendix.

\subsection{Manifold regularization}
\label{subsec:manifold}
\begin{wrapfigure}{r}{0.5\textwidth}
    \centering
    \vspace{-2mm}
    \includegraphics[width=0.4\linewidth]{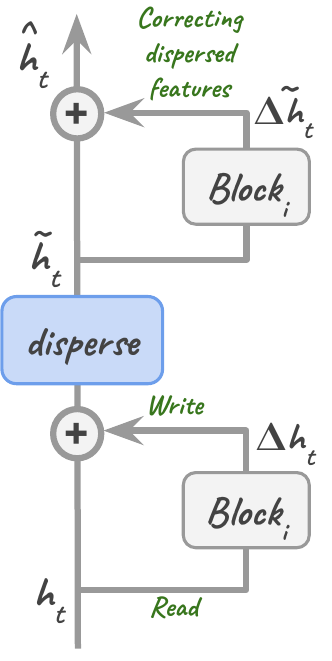}
    \caption{\textbf{Read-write mechanism of transformers:} Our regularization method leverages the read-write mechanisms of transformer blocks to add corrections to dispersed features to project them to the natural image manifold.}
    \vspace{-6mm}
    \label{fig:intuition}
\end{wrapfigure}

As with our dispersive feature update, the features $\tilde{h}_t$ can potentially be pushed out-of-the feature manifold. We propose a manifold regularization to bring the dispersed features back to the feature manifold. 
We build on the findings from seminal work in mechanistic interpretability~\cite{elhage2021mathematical} that suggest the sequence of transformer blocks with residual connections behaves in a \textit{read-write mechanism} where they read the features from the main stream ($h_t$) and add back a connection $\Delta h_t$ to the stream (see Fig~\ref{fig:intuition}). We implement the regularizer by passing the updated features $\tilde{h_t}$ again through the same MMDiT block to project them into the output feature space, giving projected features $\hat{h_t}$. In our context, we pass the dispersed features again through the same block to make corrections to $h_t$
with $\Delta \tilde{h}_t$. This allows the block, utilizing its pretrained internal priors, to 

\begin{figure}[t]
    \begin{subfigure}{0.49\textwidth}
        \centering
        \includegraphics[width=0.9\linewidth]{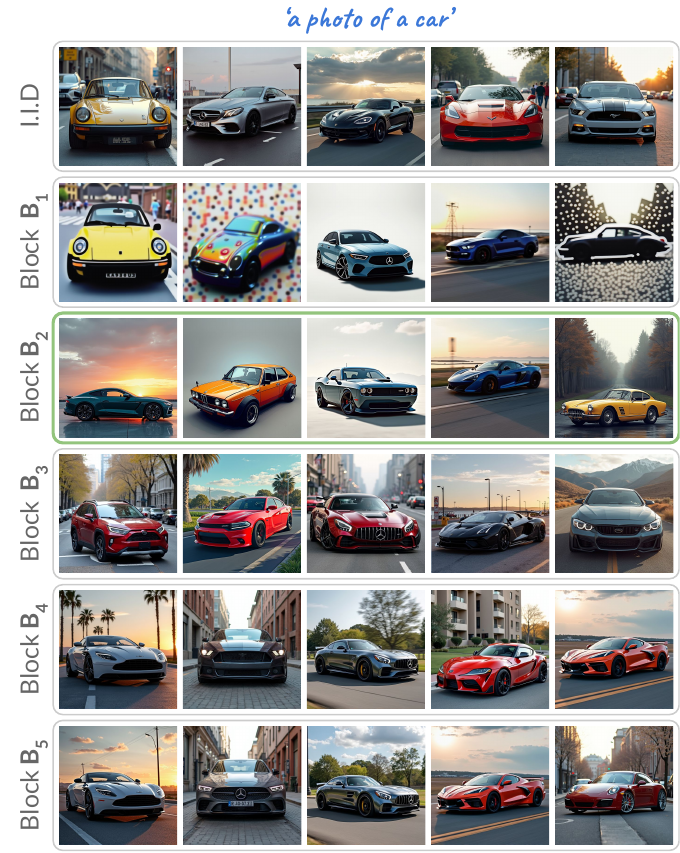}
        \caption{\textbf{Block Ablation for feature guidance:} Guidance at Block $B_2$ features is most effective, achieving high diversity while preserving fidelity. Guidance at subsequent blocks is not effective.}
        \label{fig:8_block}
    \end{subfigure}
    \hfill 
    \begin{subfigure}{0.49\textwidth}
        \centering
        \includegraphics[width=0.85\linewidth]{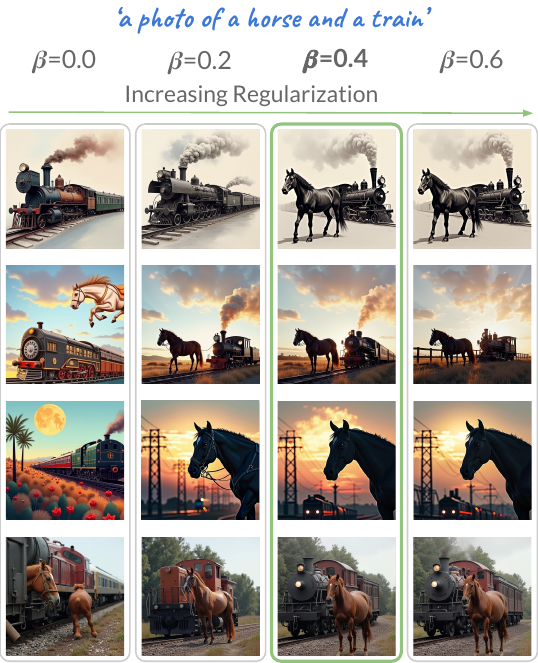}
        \caption{\textbf{Faithfulness-Diversity tradeoff:} Lower $\beta$ ensures higher diversity, but sacrifices faithfulness, while higher $\beta$ regularizes features and improves prompt adherence with a marginal diversity loss. We use $\beta = 0.4$, which offers a good tradeoff between prompt adherence and diversity.}
        \label{fig:11_qual-div}
    \end{subfigure}
    \caption{Ablation over DiT Block for dispersion \& mixing parameter $\beta$ choices.} 
    \vspace{-6mm}
\end{figure}

\noindent 
project the perturbed features back onto the natural image manifold while preserving diversity. Next, we use the projected features and the dispersed features to obtain a final update:

\begin{equation}
    h'_t := \beta \hat{h_t} + (1-\beta)\tilde{h_t}
    \vspace{-4mm}
\end{equation} 

\noindent 
where, $\beta \in [0,0.6]$. We pass the updated $h'_t$ to the next MMDiT blocks for processing. Further, $\beta$ acts as a tunable knob to steer the generation to achieve a desired faithfulness-diversity tradeoff as shown in Fig.~\ref{fig:11_qual-div}. The proposed method is a plug-and-play method that modifies outputs of only one MMDiT block and improves diversity while preserving the alignment with input conditioning.

\begin{algorithm}[H]
\caption{Diverse Generation via Feature Self-Guidance }
\label{alg:feature-guidance}
\begin{algorithmic}
    \State \textbf{Inputs:} Text embedding $c$, batch size $N$, optimization steps $N_{opt}$, learning rate $w$, temperature $\tau$, start time $T_a$, end time $T_b$, blending rate $\beta$, Blocks $\mathbf{B}_1$ to $\mathbf{B}_{57}$
    \State \textbf{Outputs:} $\{x_0^i\}_{i=1}^N$, a batch of diverse images.
\end{algorithmic}
\begin{python}
def feature_self_guidance($c$, $N$, $N_{opt}$, $w$, $\tau$, $T_a$, $T_b$, $\beta$):
    z = torch.randn(N, ., ., .) # Initial Noise
    for t in reversed(range(T)):
        if $T_a$ <= t <= $T_b$:
            $h_t = \mathbf{B}_{2} \circ \mathbf{B}_{1}(z_t, t, c)$
            for k in range($N_{opt}$):   $\rightarrow$ Self-guidance
                $\mathcal{G}$ = $\cref{eqn:1}$ ($h_t$, $\tau$)
                $h_t = h_t - w \nabla\mathcal{G}$ $\rightarrow$ Update step 
            $\hat{h_t}$ = $\mathbf{B}_2$($h_t$, t, c) $\rightarrow$ Refine 
            $h'_t = \beta \hat{h_t} + (1 - \beta) h_t $
            $v_t = \mathbf{B}_{57} \circ \cdots \mathbf{B}_{3} (h'_t, t, c)$
        else:
            $v_t = \mathbf{B}_{57}\circ \cdots \mathbf{B}_{1}(z_t, t, c)$
        $z_t = z_t + \Delta t * v_t$
    return $Decode(z_0)$
\end{python}
\end{algorithm} 
\section{Experiments}
We evaluate our method across several conditional image generation flow models - text-to-image generation (FLUX.1-dev), depth-to-image generation (FLUX.1-Depth-dev), and foundational instruction-driven image editing model (FLUX.1-Kontext-dev) for personalization. Additionally, we also evaluate the step-distilled FLUX.2-Klein-4B model, UNet-based DDPM~\cite{ddpm} and other T2I models~\cite{sana,qwen-image} in the appendix. In the following sections, we discuss the baselines, implementation details, results, and ablations.


\begin{figure*}[t]
  \centering
   \includegraphics[width=1.0\linewidth]{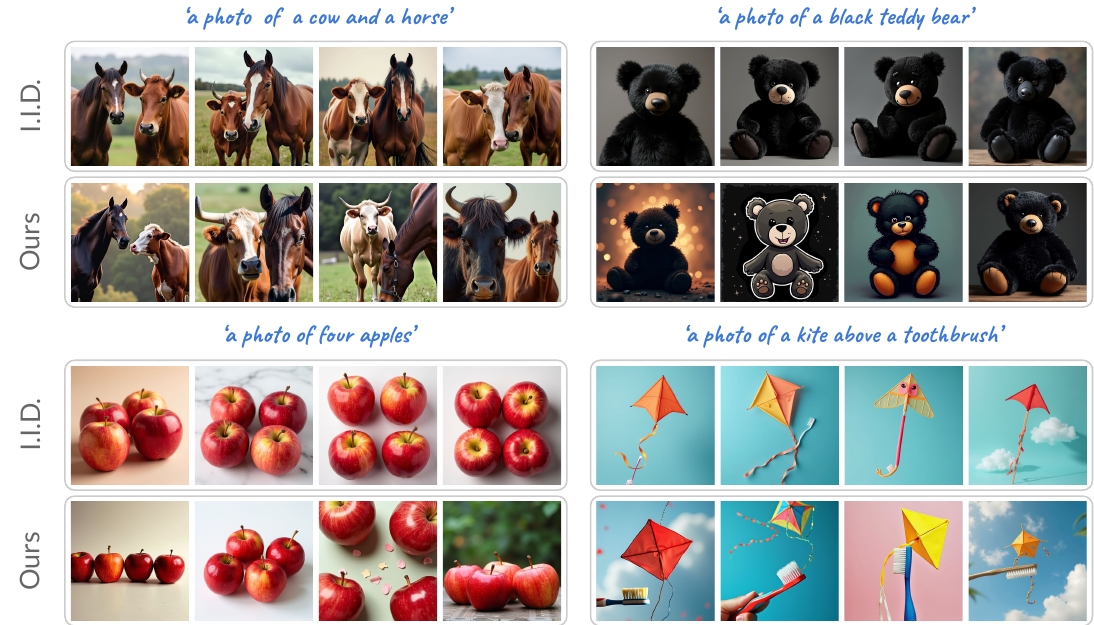}
   \caption{\textbf{Qualitative results for text-to-image generation}: We demonstrate the advantages of our inference-time method over standard I.I.D. sampling, which often suffers from visual homogeneity in layout and color. In contrast, our approach generates significant diversity in viewpoints, compositions, and color palettes while maintaining strict prompt fidelity. This is seen in the ability of our method to adhere to prompts with numerical and positional constraints without sacrificing diversity.}
   \label{fig:12_t2i_qual_results}
\end{figure*}

\label{subsec:setup}
\subsection{Experimental Setup}
\textbf{Baselines:} We compare our method with several inference approaches designed for diverse text-to-image generation. \textbf{Particle Guidance} and \textbf{Interval Guidance} leverage diffusion guidance to steer samples toward greater diversity. \textbf{Shield} \textbf{-ed} \textbf{Diffusion} implements a repellency loss that pushes the samples away from a set of shielded samples. \textbf{Group Inference} leverages pretrained CLIP~\cite{clip} and DINO~\cite{dinov2} reward models to select a subset of the most diverse and faithful samples from a large set of candidates during the denoising process. We report results for $8$ samples using $128$ and $64$ candidates for Group Inference. Furthermore, we only compare with Group Inference on depth-to-image and personalized image generation tasks, as it can be easily adapted. 

\noindent 
\textbf{Dataset:} We use task-specific evaluation datasets as discussed below:
\begin{itemize}
    \item \textbf{Text-to-Image Generation}: We use the standard \textbf{GenEval} dataset~\cite{Geneval}, an object-focused benchmark to evaluate compositional properties of text-to-image models such as object co-occurrence, position, count, and color.
    \item \textbf{Depth-Conditioned Generation}: We use a $500$ image subset of the \textbf{COCO} 2017 validation split~\cite{Coco} and extract their depth maps with DepthAnythingv2~\cite{DAv2} model for depth-to-image generation.
    \item \textbf{Personalized Image Generation}: We assess personalization results on the \textbf{Dreambooth} dataset~\cite{Dreambooth} to measure our methods' ability to generate diverse samples for a given subject with the same prompt.
\end{itemize}

\noindent\textbf{Evaluation Metrics:} We evaluate our method across two major axes: \textbf{i)} \textit{sample diversity} and \textbf{ii)} \textit{prompt adherence}. For sample diversity, we use reference-free diversity metrics: the pairwise cosine distance between DINOv2~\cite{dinov2} features, image dissimilarity using DreamSim~\cite{dreamsim}, mean similarity using MSS~\cite{cads}, VendiScore~\cite{vendiscore}, which measures the effective number of unique elements in a set and for prompt adherence, we use CLIPScore~\cite{clip,clipscore}. We report the mean and standard deviation of all the metrics across the entire dataset. We include additional metrics in the appendix.

\begin{figure*}[t]
  \centering
   \includegraphics[width=1.0\linewidth]{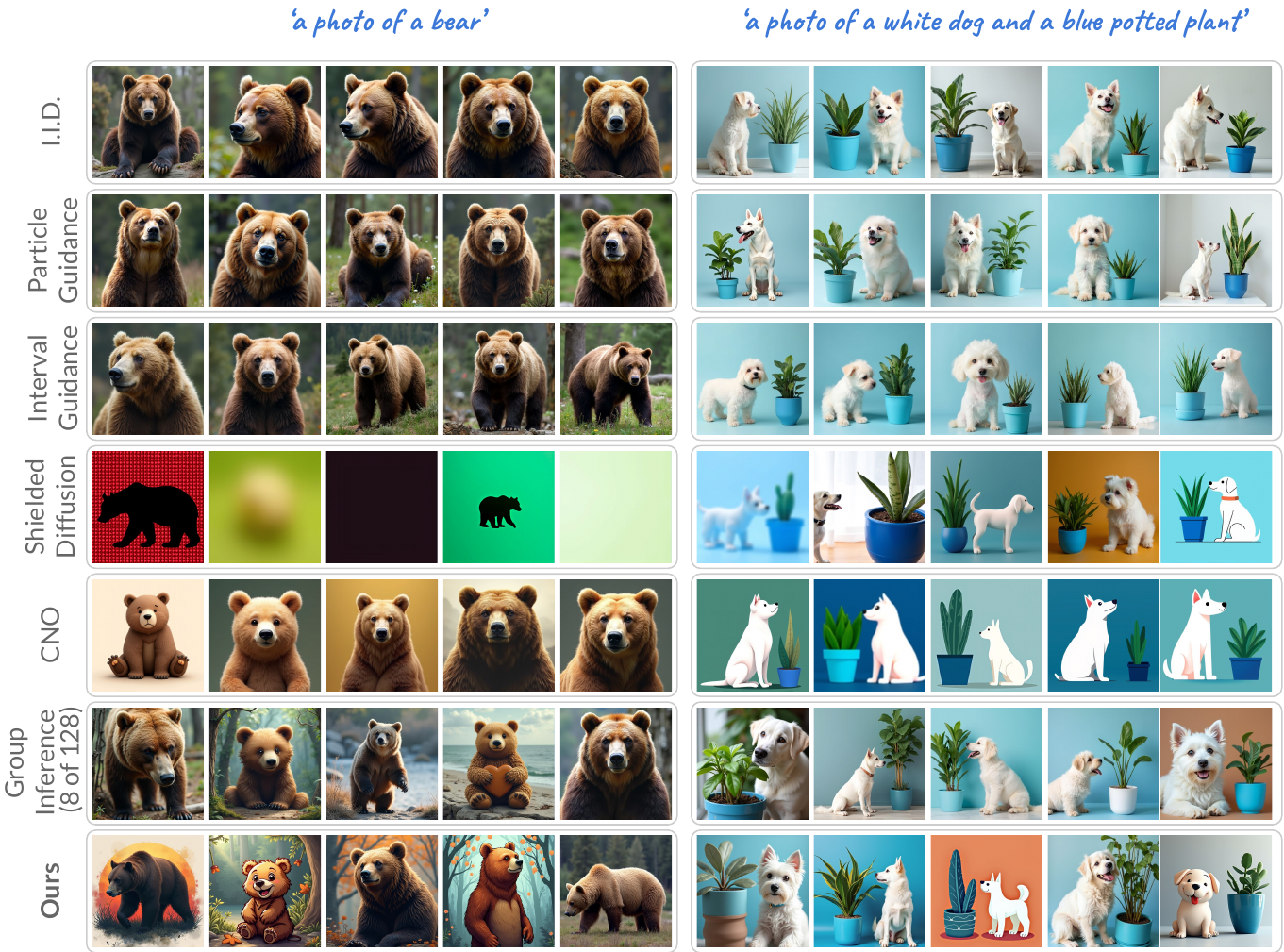}
   \caption{\textbf{Qualitative Comparison for text-to-image generation}: We compare against works on diverse sampling in T2I Generation: \textit{Particle Guidance}~\cite{particle-guidance}, \textit{Interval Guidance}~\cite{interval-guidance}, \textit{Shielded Diffusion}~\cite{shielded-diff}, \textit{CNO}~\cite{CNO} and \textit{Group Inference}~\cite{group-inference}. Our method generates highly diverse samples in terms of pose, style, and appearance while adhering to the text prompt. The baseline methods either have limited diversity (Particle Guidance, Interval Guidance, CNO) or collapse the generation (Shielded Diffusion). Group-Inference achieves good diversity; however, it requires sampling from a large number of candidates, making it much slower than our method.} 
   \vspace{-4mm}
   \label{fig:3_t2i_quals}
\end{figure*}

\vspace{2mm}
\noindent\textbf{Implementation Details:} To ensure reproducibility, all methods are evaluated using a fixed seed of $42$. We sample $8$ initial noise latents from an isotropic Gaussian distribution, $\mathcal{N}(0,I)$, and generate images at a resolution of $512^2$. For the base models, we adhere to default configurations - specifically employing a guidance scale of $3.5$ and $28$ inference steps. Our feature self-guidance is applied at the second MMDiT block during the initial denoising phase ($t \in [1,0.8]$). This process involves $30$ optimization steps with a learning rate of $0.5$ and temperature of $0.5$. Finally, the dispersed and refined latents are integrated using a blending scale of $0.4$. We use the default hyperparameters provided by baselines wherever available; a full list of hyperparameters is available in the appendix. 

\subsection{Text-to-Image Generation}
\label{subsec:results}

\noindent\textbf{Qualitative Comparisons:} We present qualitative results of our method in Fig.\ref{fig:12_t2i_qual_results}. The samples generated by our method are significantly more diverse in terms of style, layout, and view-points as compared to I.I.D. baseline. We present a comparison with several baselines for text-to-image generations in Fig.\ref{fig:3_t2i_quals}. Our method synthesizes diverse images and backgrounds from minimal textual input (e.g.\textit{`a photo of a bear'}). Our approach excels in complex compositional examples, such as \textit{`a photo of a white dog and a blue potted plant'} where it maintains distinct object identities with accurate attribute binding as it works on semantic latent features rather than spatial latents. In contrast, guidance-based methods such as Particle Guidance~\cite{particle-guidance} and Interval Guidance~\cite{interval-guidance} generate images that are visually similar to the I.I.D. baseline. Sparse intervention methods like Shielded Diffusion~\cite{shielded-diff} apply sparse repulsion to approximate clean latents, thereby pushing samples off the manifold and leading to visual artifacts (bear example). CNO~\cite{CNO} utilizes initial noise optimization to explore stylistic variations, but remains strictly tethered to the diffusion prior, limiting its output to superficial stylistic diversity rather than meaningful structural variety. Group Inference~\cite{group-inference} offers the most diversity among the baselines in terms of poses and backgrounds. However, it selects $8$ images from a candidate pool of $64/128$ images, making it inefficient for practical scenarios.

\begin{table}[t]
\centering
\caption{Quantitative results of text-to-image diverse samplers with FLUX.1-dev. }
\resizebox{\textwidth}{!}{%
\begin{tabular}{c | c | cccc | c }
\hline
Method & Latency$\downarrow$ & DINO$\uparrow$ & VS$\uparrow$ & DreamSim$\uparrow$ & MSS$\downarrow$ & CLIPScore$\uparrow$   \\ \hline

IID & \textbf{1.59s} & 0.57\tiny{$\pm$0.10}  &  2.77\tiny{$\pm$1.06} &  0.27\tiny{$\pm$0.10} & 0.32\tiny{$\pm$0.10} & 32.43\tiny{$\pm$3.71}  \\
\rowcolor[HTML]{D6D6D6} 

Particle Guidance~\cite{particle-guidance} & 2.94s & 0.58\tiny{$\pm$0.11}  & 3.16\tiny{$\pm$1.17} & 0.31\tiny{$\pm0.12$} & 0.29\tiny{$\pm$0.11} & 32.22\tiny{$\pm$3.90}  \\

Interval Guidance~\cite{interval-guidance} & 1.59s & 0.57\tiny{$\pm$0.10} & 2.88\tiny{$\pm$1.11} & 0.28\tiny{$\pm$0.10} & 0.300\tiny{$\pm$0.10} & 32.46\tiny{$\pm$3.75} \\

\rowcolor[HTML]{D6D6D6} 
Shielded Diffusion~\cite{shielded-diff} & 1.59s & 0.61\tiny{$\pm$0.10} & 3.35\tiny{$\pm$1.22} & 0.40\tiny{$\pm$0.13} & 0.23\tiny{$\pm$0.08} & 32.11\tiny{$\pm$3.89} \\

CNO~\cite{CNO} & 1.75s & 0.63\tiny{$\pm$0.08} & 3.46\tiny{$\pm$1.10} & 0.39\tiny{$\pm$0.10} & 0.22\tiny{$\pm$0.07} & 31.97\tiny{$\pm$3.77} \\

\rowcolor[HTML]{D6D6D6}  
GroupInference~\cite{group-inference} (8 of 64) & 2.86s & 0.67\tiny{$\pm$0.07} & 3.37\tiny{$\pm$1.17} & 0.35\tiny{$\pm$0.10} & 0.21\tiny{$\pm$0.07} & \textbf{32.49\tiny{$\pm$3.74}} \\

\rowcolor[HTML]{D6D6D6}  
GroupInference~\cite{group-inference} (8 of 128) & 4.55s & \textbf{0.70\tiny{$\pm$0.07}} & \textbf{3.61\tiny{$\pm$1.26}} & 0.37\tiny{$\pm$0.11} & 0.19\tiny{$\pm$0.07} & 32.37\tiny{$\pm$3.77} \\ \hline

\rowcolor[HTML]{90EE90} 
Ours & 1.70s & 0.68\tiny{$\pm$0.07} & 3.57\tiny{$\pm$1.17} & \textbf{0.40\tiny{$\pm$0.09}} & \textbf{0.19\tiny{$\pm$0.06}} & 32.05\tiny{$\pm$3.64} \\ \hline
\end{tabular}
}
\label{tab:table1}
\end{table}

\vspace{2mm}
\noindent\textbf{Quantitative Comparisons:} We present our quantitative results in Tab.\ref{tab:table1}. Analysis of the metrics reveals several limitations in existing state-of-the-art methods. Both Particle Guidance \cite{particle-guidance} and Interval Guidance \cite{interval-guidance} fail to surpass the I.I.D. baseline in generative diversity; furthermore, Particle Guidance is an inefficient paradigm due to its prohibitive memory requirements as seen in Fig.\ref{fig:8_compute} \textbf{b)}. Shielded Diffusion \cite{shielded-diff} although time efficient, is limited in the sample diversity and has inferior overall prompt adherence. Similarly, CNO \cite{CNO} is limited in its diversity to stylistic variations and is unable to capture all the modes of the distribution. Finally, while Group Inference \cite{group-inference} obtains high quality and diversity, it is extremely slow due to the selection of a subset from a large pool, as reported in latency. Its reliance on additional reward models to score the large pool of samples increases inference time, making it less practical for real-time applications. In contrast, our method offers significantly enhanced diversity while preserving prompt adherence and comparable latency to I.I.D. generation, making it highly practical for real-world scenarios. These results underscore our method's ability to unlock structural and stylistic diversity without compromising the underlying diffusion prior's stability or usability. We compare with additional baselines and also evaluate on DPG-Bench~\cite{dpgbench} in the appendix.

\setlength{\intextsep}{2pt}
\setlength{\columnsep}{5pt}
\begin{figure*}[t]
  \centering
    \begin{subfigure}{0.495\textwidth}
        \includegraphics[width=\linewidth]{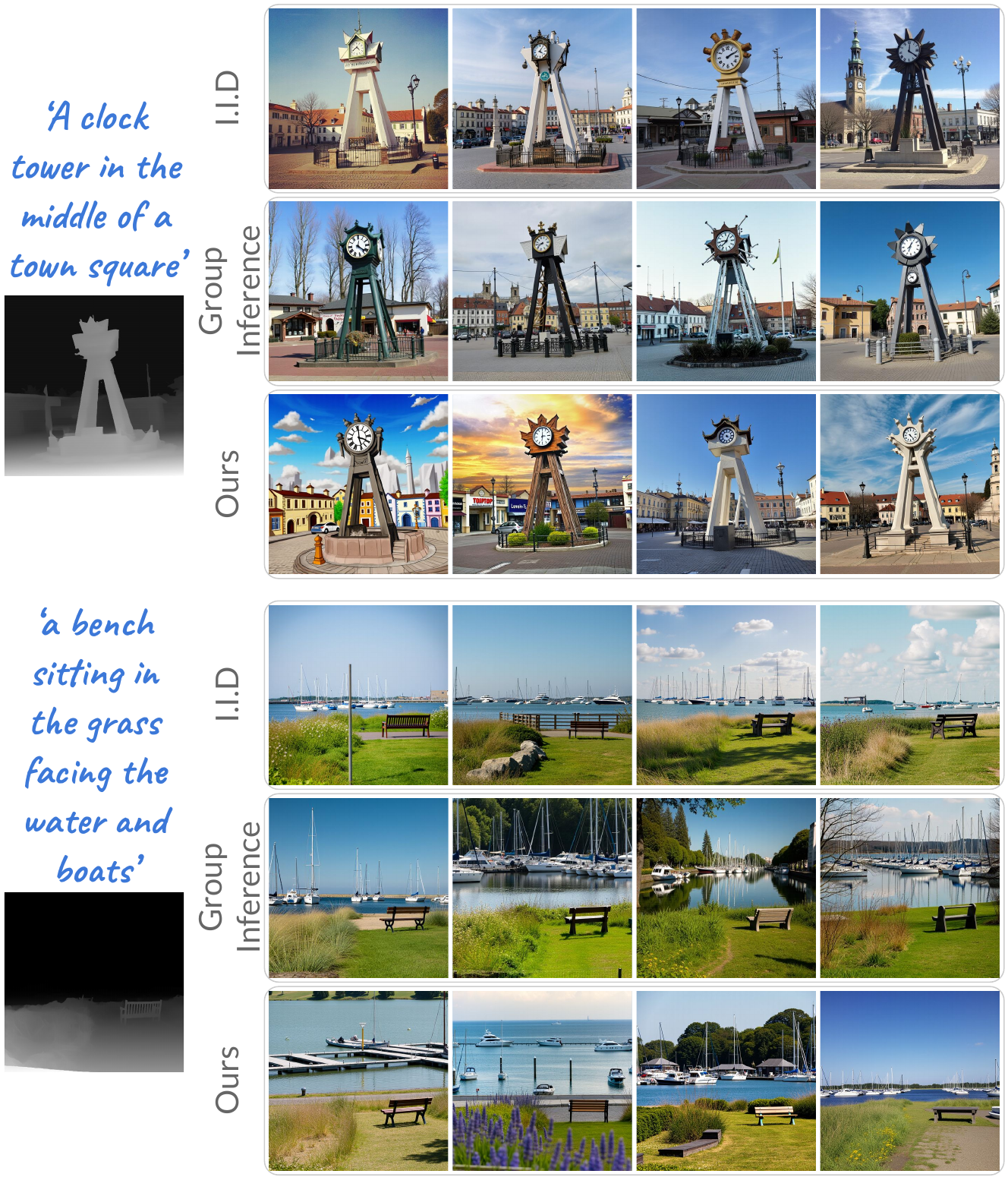}
        \caption{Depth-to-Image Generation}
        \label{fig:4_d2i_quals}
    \end{subfigure}
    \hfill
    \begin{subfigure}{0.495\textwidth}
        \includegraphics[width=\linewidth]{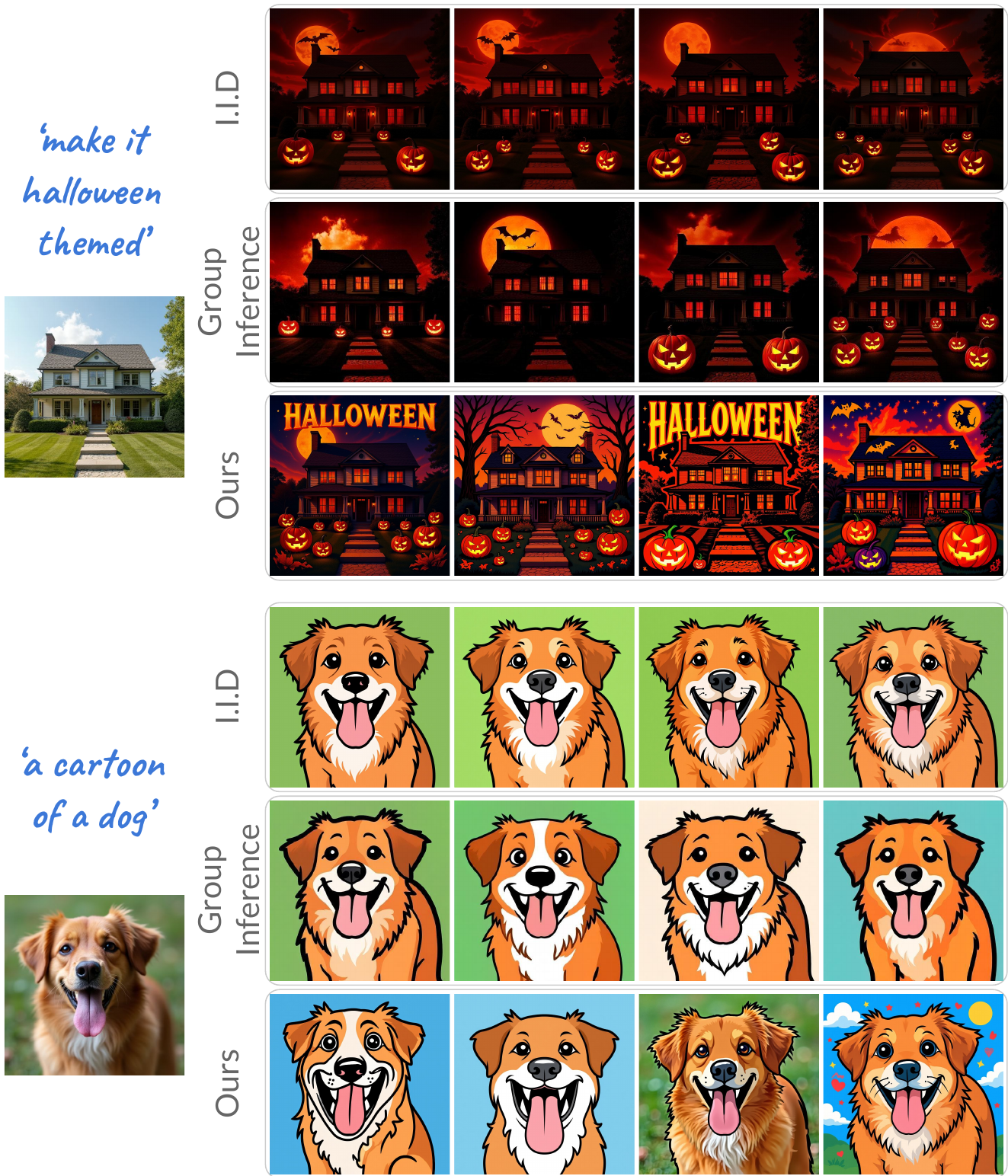}
        \caption{Reference-Image Generation}
        \label{fig:5_p2i_quals}
    \end{subfigure}
    \caption{\textbf{Additional Generation Tasks}: Our method demonstrates the ability to generate diverse samples while maintaining adherence to the input conditioning, such as \textbf{depth maps} or \textbf{reference images}. For depth-conditioned generation, our method generates diverse backgrounds and textures. For reference conditioned generation, our method preserves the house/dog identity and generates diverse styles, whereas baseline methods have limited sample diversity.}
    \vspace{-4mm}
    \label{fig:4_5_conditional_quals}
\end{figure*}

\noindent 

\subsection{Additional Image Generation Tasks}
\label{subsec:condition-gen}

\begin{wraptable}{r}{0.5\textwidth}
\vspace{-4mm}
\centering
\caption{\textbf{Quantitative results of conditional image generation:} Our method significantly improves diversity over I.I.D. while preserving the prompt alignment.}
\label{tab:2_conditional}
\resizebox{0.5\textwidth}{!}{
\begin{tabular}{ccc|cc}
\hline
Task & \multicolumn{2}{c|}{\textbf{Depth to Image}} & \multicolumn{2}{l}{\textbf{Personalized Image}} \\ \hline
Method$\downarrow$ & DINO & CLIPScore & DINO & CLIPScore \\ \hline  
I.I.D. & 0.43\tiny{$\pm$0.08}  & \textbf{30.85\tiny{$\pm$3.12}}  & 0.45\tiny{$\pm$0.10} & 31.93\tiny{$\pm$4.31} \\
Group Inference & 0.47\tiny{$\pm$0.09} & 30.63\tiny{$\pm$3.12} & 0.49\tiny{$\pm$0.15} & \textbf{32.58\tiny{$\pm$3.54}} \\
\rowcolor[HTML]{90EE90} 
Ours & \textbf{0.52\tiny{$\pm$0.09}} & 30.49\tiny{$\pm$3.23} & \textbf{0.57\tiny{$\pm$0.09}} & 32.04\tiny{$\pm$4.11} \\ \hline
\end{tabular}
}
\end{wraptable}

We show the generalization of our method by integrating it with other conditional flow models. We demonstrate results on depth-conditioned generation and reference-conditioned generation, where the same subject needs to be generated in diverse contexts in Fig.\ref{fig:4_5_conditional_quals} and Tab.\ref{tab:2_conditional}. Our method consistently produces diverse compositions and styles, and faithfully adheres to strong conditioning like depth maps (Fig.\ref{fig:4_d2i_quals}) and reference images (Fig.\ref{fig:5_p2i_quals}). 
In contrast, Group Inference has inferior diversity and has high inference latency. 
These results are further quantified in  Tab.\ref{tab:2_conditional}, which demonstrates our method's superior diversity with marginal impact on overall prompt adherence. 

\label{subsec:ablations}
\subsection{Ablations} 

\noindent
We conduct ablation studies to evaluate the key design choices, including the regularization parameter $\beta$ and inference batch size. Further, we provide the ablation over the selection of MMDiT block and the specific denoising timestep interval for feature self-guidance in the appendix. 

\vspace{2mm}
\noindent\textbf{Faithfulness-Diversity Tradeoff with $\beta$:} The regularization parameter $\beta$ allows for a trade-off between the faithfulness, i.e., prompt adherence, and diversity of samples. We illustrate this in Fig.\ref{fig:11_qual-div} and Fig.\ref{fig:8_compute} \textbf{a)}, where lower $\beta$ has higher diversity but inferior prompt-adherence and vice-versa. This provides a smooth control to the user to adapt our method based on the final use-case. We evaluate this tradeoff across baselines in the appendix.

\begin{wrapfigure}[]{r}{0.5\textwidth}
\centering
    \includegraphics[width=\linewidth]{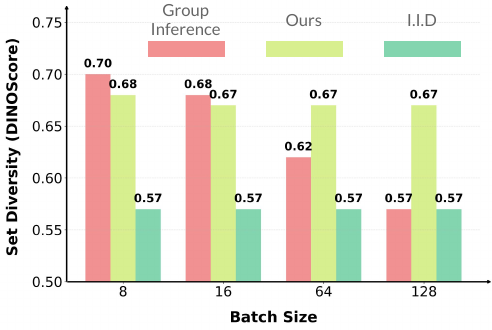}
    \vspace{-4mm}
    \caption{Performance on scaling Batch Size}
    \label{fig:9_batch}
\end{wrapfigure}

\vspace{2mm}
\noindent\textbf{Scaling Batch Size}: To evaluate the robustness of our approach for different batch sizes, we ablate the generation with batch size upto $128$ samples. As illustrated in Fig.\ref{fig:9_batch}, our method maintains high diversity and prompt alignment even at elevated scales, demonstrating superior scalability. In contrast, Group Inference~\cite{group-inference} exhibits diminishing diversity as the batch size increases. As the fraction of selected samples increases, inability to discard the similar samples causes diversity to regress toward I.I.D. baseline performance. Our method avoids these sampling bottlenecks, preserving output faithfulness and variance across large-scale generations.

\subsection{Computation Analysis}
\label{subsec:computation}

\begin{figure}
\centering
    \includegraphics[width=\linewidth]{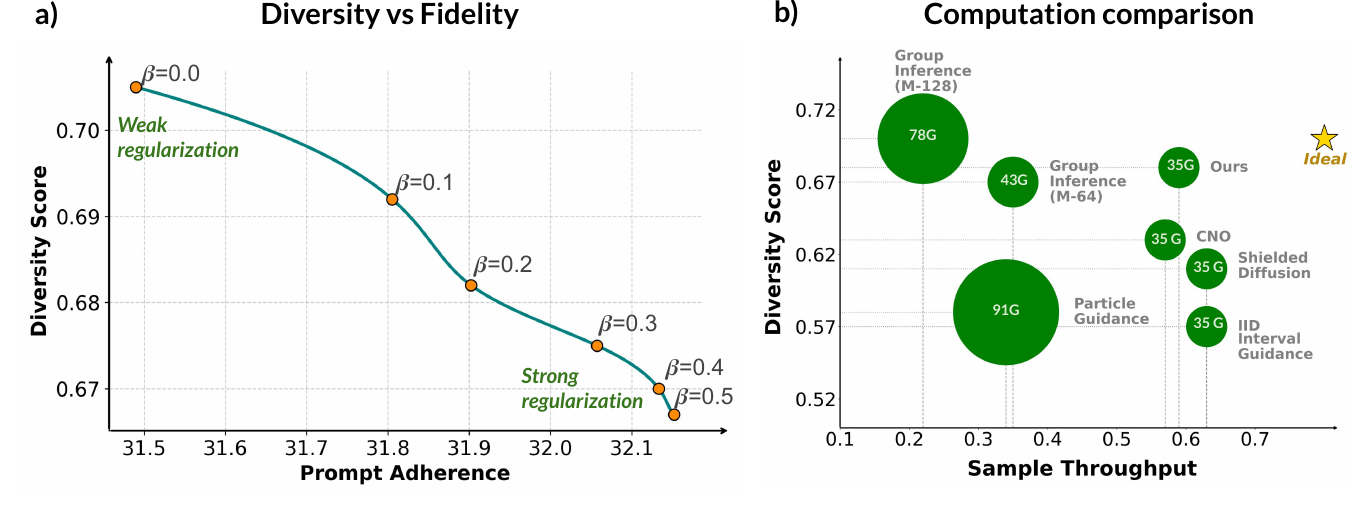}
    \caption{\textbf{a) Faithfulness to the Prompt and Diversity Pareto front for our method.} Varying $\beta$ offers a tradeoff between prompt adherence and diversity, higher beta promotes stronger regularization and improves prompt adherence and vice-versa. \textbf{b) Computational Cost v/s. Diversity on FLUX.1-dev}: We show a breakdown of the memory consumption, sample throughput and diversity of all methods. Radius of the bubbles indicate VRAM consumption. Our method achieves high diversity alongwith good sample throughput.}
    \label{fig:8_compute}
\end{figure}

\noindent We evaluate the efficiency and computational overhead of our proposed method relative to the baselines. All the benchmarking experiments were conducted using the FLUX.1-dev \cite{flux} model on a single NVIDIA H200 GPU. Our evaluation specifically characterizes the trade-off between enhanced generative diversity and incremental computational cost in terms of sample throughput and GPU VRAM used. To ensure statistical reliability, we generate $400$ samples per method, incorporating a preliminary warmup phase, and record the average sample throughput and peak VRAM utilization. We present the comparison in Fig.\ref{fig:8_compute} \textbf{b)}. In this visualization, bubble size 
is proportionate to the VRAM utilization, the x-axis is the sample throughput, which is the number of samples generated per second, and the y-axis is the DINO diversity score. Our method achieves high diversity with a computational footprint nearly identical to the I.I.D. baseline. Group Inference has marginally higher diversity than ours at a cost of smaller throughput and more VRAM, limiting its practical utility. All the other baselines, though efficient, do not achieve good diversity. Our method achieves an excellent tradeoff in terms of quality, efficiency, and memory required.

\subsection{User Study}
\label{subsec:user-study}

\begin{wrapfigure}{r}{0.5\textwidth}
    \vspace{-7mm}
    \centering
    \includegraphics[width=0.90\linewidth]{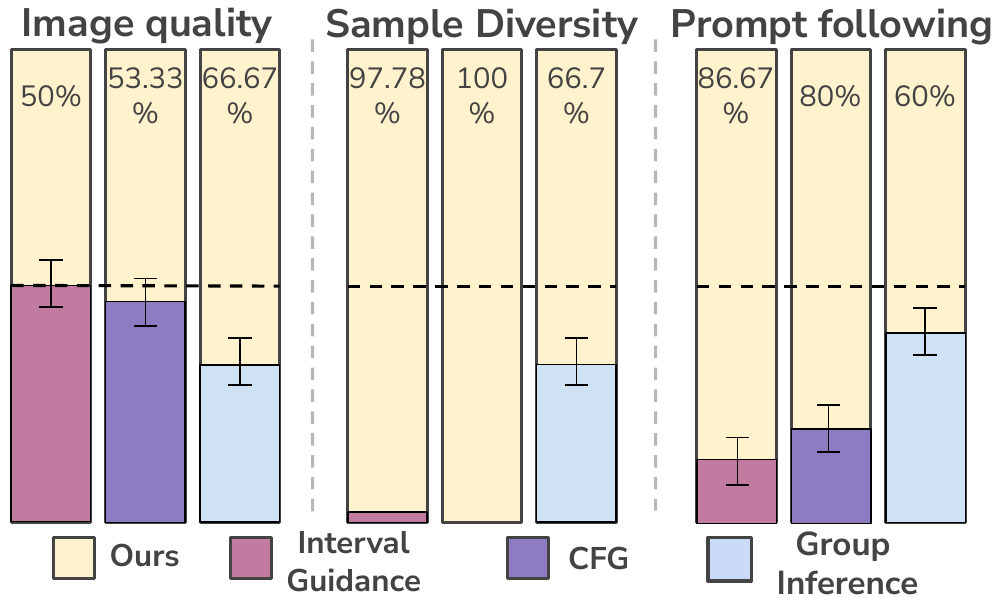}
    \caption{\textbf{User Study:} User preferences show that our method substantially outperforms baselines in sample diversity while delivering competitive image quality and prompt following scores.}
    \label{appendix-fig:user-study}
    \vspace{-6mm}
\end{wrapfigure}

We conduct a user study to subjectively evaluate the samples generated from our method against three baselines \textbf{a)} Interval Guidance, \textbf{b)} CFG and \textbf{c)} Group Inference across three dimensions: \textbf{i)} \textit{Image Quality}, \textbf{ii)} \textit{Prompt Alignment} and \textbf{iii)} \textit{Image Diversity}. The study consists of 20 comparison pairs evaluated by 15 volunteers. Results of the user study are summarized in Fig.~\ref{appendix-fig:user-study}

\vspace{1mm}
\section{Conclusion and Discussion}
\noindent
\textbf{Limitations.} While our approach excels in effectively improving diversity, it has some limitations.  As an inference-time intervention, our method is inherently limited by the base model and may inherit biases present in the pretrained model. Additionally, because the approach relies on batch-level computations, it is not effective in streaming image generation. 

\vspace{2mm}
\noindent
\textbf{Conclusion.} We introduce a training-free framework designed to systematically rectify diversity collapse in pretrained flow models. Our methodology is built on the key insight that diversity collapse stems from the collapse of internal features of the flow model. To address this, we present a lightweight feature self-guidance mechanism that effectively disperses these features, coupled with a manifold regularization step to ensure that updated features remain anchored to the data distribution. This two-step approach maintains high structural fidelity and alignment with input prompts while functioning as a seamless, plug-and-play module across various conditional flow generative models. Our method takes an important step towards efficiently and effectively improving the pretrained flow models and can fuel more research in investigating the internal features of the flow models. As a future direction, we aim to extend this framework to enhance diversity in other modalities such as 3D model synthesis, video generation, and the generation of diverse molecular structures.

\section*{Acknowledgement}

We thank Tejan Karmali and Vaibhav Agrawal for reviewing the manuscript. This work is supported by PMRF by the Govt. of India. 

\bibliographystyle{splncs04}
\bibliography{main}

@String(ICCV  = {Int. Conf. Comput. Vis.})

@String(TOG   = {ACM Trans. Graph.})

@String(ICCV  = {ICCV})

@String(TOG   = {ACM TOG})

@inproceedings{sd3,
  title={Scaling rectified flow transformers for high-resolution image synthesis},
  author={Esser, Patrick and Kulal, Sumith and Blattmann, Andreas and Entezari, Rahim and M{\"u}ller, Jonas and Saini, Harry and Levi, Yam and Lorenz, Dominik and Sauer, Axel and Boesel, Frederic and others},
  booktitle={Forty-first international conference on machine learning},
  year={2024}
}

@misc{flux,
    author={Black Forest Labs},
    title={FLUX},
    year={2024},
    howpublished={\url{https://github.com/black-forest-labs/flux}},
}

@misc{flux-2,
    author={Black Forest Labs},
    title={{FLUX.2: Frontier Visual Intelligence}},
    year={2025},
    howpublished={\url{https://bfl.ai/blog/flux-2}},
}

@article{qwen-image,
  title={Qwen-image technical report},
  author={Wu, Chenfei and Li, Jiahao and Zhou, Jingren and Lin, Junyang and Gao, Kaiyuan and Yan, Kun and Yin, Sheng-ming and Bai, Shuai and Xu, Xiao and Chen, Yilei and others},
  journal={arXiv preprint arXiv:2508.02324},
  year={2025}
}

@article{sana,
  title={Sana 1.5: Efficient scaling of training-time and inference-time compute in linear diffusion transformer},
  author={Xie, Enze and Chen, Junsong and Zhao, Yuyang and Yu, Jincheng and Zhu, Ligeng and Wu, Chengyue and Lin, Yujun and Zhang, Zhekai and Li, Muyang and Chen, Junyu and others},
  journal={arXiv preprint arXiv:2501.18427},
  year={2025}
}

@article{repa,
  title={Representation alignment for generation: Training diffusion transformers is easier than you think},
  author={Yu, Sihyun and Kwak, Sangkyung and Jang, Huiwon and Jeong, Jongheon and Huang, Jonathan and Shin, Jinwoo and Xie, Saining},
  journal={arXiv preprint arXiv:2410.06940},
  year={2024}
}

@article{rae,
  title={Diffusion transformers with representation autoencoders},
  author={Zheng, Boyang and Ma, Nanye and Tong, Shengbang and Xie, Saining},
  journal={arXiv preprint arXiv:2510.11690},
  year={2025}
}

@article{irepa,
  title={What matters for Representation Alignment: Global Information or Spatial Structure?},
  author={Singh, Jaskirat and Leng, Xingjian and Wu, Zongze and Zheng, Liang and Zhang, Richard and Shechtman, Eli and Xie, Saining},
  journal={arXiv preprint arXiv:2512.10794},
  year={2025}
}

@article{scaling-rae,
  title={Scaling Text-to-Image Diffusion Transformers with Representation Autoencoders},
  author={Tong, Shengbang and Zheng, Boyang and Wang, Ziteng and Tang, Bingda and Ma, Nanye and Brown, Ellis and Yang, Jihan and Fergus, Rob and LeCun, Yann and Xie, Saining},
  journal={arXiv preprint arXiv:2601.16208},
  year={2026}
}

@article{diffuse-disperse,
  title={Diffuse and Disperse: Image Generation with Representation Regularization},
  author={Wang, Runqian and He, Kaiming},
  journal={arXiv preprint arXiv:2506.09027},
  year={2025}
}

@article{internal-guidance,
  title={Guiding a Diffusion Transformer with the Internal Dynamics of Itself},
  author={Zhou, Xingyu and Li, Qifan and Hu, Xiaobin and Chen, Hai and Gu, Shuhang},
  journal={arXiv preprint arXiv:2512.24176},
  year={2025}
}

@inproceedings{bansal2023universal,
  title={Universal guidance for diffusion models},
  author={Bansal, Arpit and Chu, Hong-Min and Schwarzschild, Avi and Sengupta, Soumyadip and Goldblum, Micah and Geiping, Jonas and Goldstein, Tom},
  booktitle={Proceedings of the IEEE/CVF conference on computer vision and pattern recognition},
  pages={843--852},
  year={2023}
}

@inproceedings{voynov2023sketch,
  title={Sketch-guided text-to-image diffusion models},
  author={Voynov, Andrey and Aberman, Kfir and Cohen-Or, Daniel},
  booktitle={ACM SIGGRAPH 2023 conference proceedings},
  pages={1--11},
  year={2023}
}

@article{epstein2023diffusion,
  title={Diffusion self-guidance for controllable image generation},
  author={Epstein, Dave and Jabri, Allan and Poole, Ben and Efros, Alexei and Holynski, Aleksander},
  journal={Advances in Neural Information Processing Systems},
  volume={36},
  pages={16222--16239},
  year={2023}
}

@article{particle-guidance,
  title={Particle guidance: non-iid diverse sampling with diffusion models},
  author={Corso, Gabriele and Xu, Yilun and De Bortoli, Valentin and Barzilay, Regina and Jaakkola, Tommi},
  journal={arXiv preprint arXiv:2310.13102},
  year={2023}
}

@article{interval-guidance,
  title={Applying guidance in a limited interval improves sample and distribution quality in diffusion models},
  author={Kynk{\"a}{\"a}nniemi, Tuomas and Aittala, Miika and Karras, Tero and Laine, Samuli and Aila, Timo and Lehtinen, Jaakko},
  journal={Advances in Neural Information Processing Systems},
  volume={37},
  pages={122458--122483},
  year={2024}
}

@article{cads,
  title={CADS: Unleashing the diversity of diffusion models through condition-annealed sampling},
  author={Sadat, Seyedmorteza and Buhmann, Jakob and Bradley, Derek and Hilliges, Otmar and Weber, Romann M},
  journal={arXiv preprint arXiv:2310.17347},
  year={2023}
}

@article{alaluf2023neural,
  title={A neural space-time representation for text-to-image personalization},
  author={Alaluf, Yuval and Richardson, Elad and Metzer, Gal and Cohen-Or, Daniel},
  journal={ACM Transactions on Graphics (TOG)},
  volume={42},
  number={6},
  pages={1--10},
  year={2023},
  publisher={ACM New York, NY, USA}
}

@article{gal2023encoder,
  title={Encoder-based domain tuning for fast personalization of text-to-image models},
  author={Gal, Rinon and Arar, Moab and Atzmon, Yuval and Bermano, Amit H and Chechik, Gal and Cohen-Or, Daniel},
  journal={ACM Transactions on Graphics (TOG)},
  volume={42},
  number={4},
  pages={1--13},
  year={2023},
  publisher={ACM New York, NY, USA}
}

@inproceedings{galimage,
title={An Image is Worth One Word: Personalizing Text-to-Image Generation using Textual Inversion},
author={Rinon Gal and Yuval Alaluf and Yuval Atzmon and Or Patashnik and Amit Haim Bermano and Gal Chechik and Daniel Cohen-Or},
booktitle={The Eleventh International Conference on Learning Representations },
year={2023},
url={https://openreview.net/forum?id=NAQvF08TcyG}
}

@article{shielded-diff,
  title={Shielded Diffusion: Generating Novel and Diverse Images using Sparse Repellency},
  author={Kirchhof, Michael and Thornton, James and B{\'e}thune, Louis and Ablin, Pierre and Ndiaye, Eugene and Cuturi, Marco},
  journal={arXiv preprint arXiv:2410.06025},
  year={2024}
}

@article{CNO,
  title={Diverse Text-to-Image Generation via Contrastive Noise Optimization},
  author={Kim, Byungjun and Um, Soobin and Ye, Jong Chul},
  journal={arXiv preprint arXiv:2510.03813},
  year={2025}
}

@article{CFG,
  title={Classifier-free diffusion guidance},
  author={Ho, Jonathan and Salimans, Tim},
  journal={arXiv preprint arXiv:2207.12598},
  year={2022}
}

@article{classifier-guidance,
  title={Diffusion models beat gans on image synthesis},
  author={Dhariwal, Prafulla and Nichol, Alexander},
  journal={Advances in neural information processing systems},
  volume={34},
  pages={8780--8794},
  year={2021}
}

@inproceedings{ERG,
title={Entropy Rectifying Guidance for Diffusion and Flow Models},
author={Tariq Berrada and Adriana Romero-Soriano and Michal Drozdzal and Jakob Verbeek and Karteek Alahari},
booktitle={The Thirty-ninth Annual Conference on Neural Information Processing Systems},
year={2025},
url={https://openreview.net/forum?id=RjN1LYymST}
}

@article{Sparke,
  author = {Mohammad Jalali and Haoyu Lei and Amin Gohari and Farzan Farnia},
  title = {SPARKE: Scalable Prompt-Aware Diversity Guidance in Diffusion Models via RKE Score},
  journal = {arXiv preprint arXiv:2506.10173},
  year = {2025},
  url = {https://arxiv.org/abs/2506.10173},
}

@article{group-inference,
  title={Scaling group inference for diverse and high-quality generation},
  author={Parmar, Gaurav and Patashnik, Or and Ostashev, Daniil and Wang, Kuan-Chieh and Aberman, Kfir and Narasimhan, Srinivasa and Zhu, Jun-Yan},
  journal={arXiv preprint arXiv:2508.15773},
  year={2025}
}

@article{nevertoolate,
  title={It's Never Too Late: Noise Optimization for Collapse Recovery in Trained Diffusion Models},
  author={Harrington, Anne and Koepke, A and Karthik, Shyamgopal and Darrell, Trevor and Efros, Alexei A},
  journal={arXiv preprint arXiv:2601.00090},
  year={2025}
}

@article{dpp-grpo-video,
  title={Diverse Video Generation with Determinantal Point Process-Guided Policy Optimization},
  author={Kazimi, Tahira and Dunlop, Connor and Yanardag, Pinar},
  journal={arXiv preprint arXiv:2511.20647},
  year={2025}
}

@article{noise-hypernetworks,
  title={Noise hypernetworks: Amortizing test-time compute in diffusion models},
  author={Eyring, Luca and Karthik, Shyamgopal and Dosovitskiy, Alexey and Ruiz, Nataniel and Akata, Zeynep},
  journal={arXiv preprint arXiv:2508.09968},
  year={2025}
}

@article{DNO,
  title={Inference-time alignment of diffusion models with direct noise optimization},
  author={Tang, Zhiwei and Peng, Jiangweizhi and Tang, Jiasheng and Hong, Mingyi and Wang, Fan and Chang, Tsung-Hui},
  journal={arXiv preprint arXiv:2405.18881},
  year={2024}
}

@inproceedings{DreamDistribution,
  title={DreamDistribution: Learning prompt distribution for diverse in-distribution generation},
  author={Zhao, Brian Nlong and Xiao, Yuhang and Xu, Jiashu and Jiang, Xinyang and Yang, Yifan and Li, Dongsheng and Itti, Laurent and Vineet, Vibhav and Ge, Yunhao},
  booktitle={The Thirteenth International Conference on Learning Representations},
  year={2025}
}

@inproceedings{minority-prompt,
  title={Minority-focused text-to-image generation via prompt optimization},
  author={Um, Soobin and Ye, Jong Chul},
  booktitle={Proceedings of the Computer Vision and Pattern Recognition Conference},
  pages={20926--20936},
  year={2025}
}

@inproceedings{learning-to-sample-diverse-prompts,
  title={Learning to sample effective and diverse prompts for text-to-image generation},
  author={Yun, Taeyoung and Zhang, Dinghuai and Park, Jinkyoo and Pan, Ling},
  booktitle={Proceedings of the Computer Vision and Pattern Recognition Conference},
  pages={23625--23635},
  year={2025}
}

@InProceedings{TIR,
    author    = {Khan, Mohammed Abdul Hafeez and Jain, Yash and Bhattacharyya, Siddhartha and Vineet, Vibhav},
    title     = {Test-time Prompt Refinement for Text-to-Image Models},
    booktitle = {Proceedings of the IEEE/CVF International Conference on Computer Vision (ICCV) Workshops},
    month     = {October},
    year      = {2025},
    pages     = {6506-6516}
}

@InProceedings{ReflectionFlow,
    author    = {Zhuo, Le and Zhao, Liangbing and Paul, Sayak and Liao, Yue and Zhang, Renrui and Xin, Yi and Gao, Peng and Elhoseiny, Mohamed and Li, Hongsheng},
    title     = {From Reflection to Perfection: Scaling Inference-Time Optimization for Text-to-Image Diffusion Models via Reflection Tuning},
    booktitle = {Proceedings of the IEEE/CVF International Conference on Computer Vision (ICCV)},
    month     = {October},
    year      = {2025},
    pages     = {15329-15339}
}

@inproceedings{Dual-process,
  title={Dual-process image generation},
  author={Luo, Grace and Granskog, Jonathan and Holynski, Aleksander and Darrell, Trevor},
  booktitle={Proceedings of the IEEE/CVF International Conference on Computer Vision},
  pages={17972--17983},
  year={2025}
}

@article{Saining-inference,
  title={Inference-time scaling for diffusion models beyond scaling denoising steps},
  author={Ma, Nanye and Tong, Shangyuan and Jia, Haolin and Hu, Hexiang and Su, Yu-Chuan and Zhang, Mingda and Yang, Xuan and Li, Yandong and Jaakkola, Tommi and Jia, Xuhui and others},
  journal={arXiv preprint arXiv:2501.09732},
  year={2025}
}

@article{Minhyuk-inference,
  title={Inference-time scaling for flow models via stochastic generation and rollover budget forcing},
  author={Kim, Jaihoon and Yoon, Taehoon and Hwang, Jisung and Sung, Minhyuk},
  journal={arXiv preprint arXiv:2503.19385},
  year={2025}
}

@article{CoT,
  title={Chain-of-thought prompting elicits reasoning in large language models},
  author={Wei, Jason and Wang, Xuezhi and Schuurmans, Dale and Bosma, Maarten and Xia, Fei and Chi, Ed and Le, Quoc V and Zhou, Denny and others},
  journal={Advances in neural information processing systems},
  volume={35},
  pages={24824--24837},
  year={2022}
}

@article{llm-scaling,
  title={Scaling llm test-time compute optimally can be more effective than scaling model parameters},
  author={Snell, Charlie and Lee, Jaehoon and Xu, Kelvin and Kumar, Aviral},
  journal={arXiv preprint arXiv:2408.03314},
  year={2024}
}

@article{reno,
  title={Reno: Enhancing one-step text-to-image models through reward-based noise optimization},
  author={Eyring, Luca and Karthik, Shyamgopal and Roth, Karsten and Dosovitskiy, Alexey and Akata, Zeynep},
  journal={Advances in Neural Information Processing Systems},
  volume={37},
  pages={125487--125519},
  year={2024}
}

@article{ddpm,
  title={Denoising diffusion probabilistic models},
  author={Ho, Jonathan and Jain, Ajay and Abbeel, Pieter},
  journal={Advances in neural information processing systems},
  volume={33},
  pages={6840--6851},
  year={2020}
}

@article{ddim,
  title={Denoising diffusion implicit models},
  author={Song, Jiaming and Meng, Chenlin and Ermon, Stefano},
  journal={arXiv preprint arXiv:2010.02502},
  year={2020}
}

@article{score-sde,
  title={Score-based generative modeling through stochastic differential equations},
  author={Song, Yang and Sohl-Dickstein, Jascha and Kingma, Diederik P and Kumar, Abhishek and Ermon, Stefano and Poole, Ben},
  journal={arXiv preprint arXiv:2011.13456},
  year={2020}
}

@article{flow-matching,
  title={Flow matching for generative modeling},
  author={Lipman, Yaron and Chen, Ricky TQ and Ben-Hamu, Heli and Nickel, Maximilian and Le, Matt},
  journal={arXiv preprint arXiv:2210.02747},
  year={2022}
}

@article{rectified-flow,
  title={Flow straight and fast: Learning to generate and transfer data with rectified flow},
  author={Liu, Xingchao and Gong, Chengyue and Liu, Qiang},
  journal={arXiv preprint arXiv:2209.03003},
  year={2022}
}

@article{h-space,
  title={Diffusion models already have a semantic latent space},
  author={Kwon, Mingi and Jeong, Jaeseok and Uh, Youngjung},
  journal={arXiv preprint arXiv:2210.10960},
  year={2022}
}

@inproceedings{balancing-act,
  title={Balancing act: Distribution-guided debiasing in diffusion models},
  author={Parihar, Rishubh and Bhat, Abhijnya and Basu, Abhipsa and Mallick, Saswat and Kundu, Jogendra Nath and Babu, R Venkatesh},
  booktitle={Proceedings of the IEEE/CVF conference on computer vision and pattern recognition},
  pages={6668--6678},
  year={2024}
}

@inproceedings{StableFlow,
  title={Stable flow: Vital layers for training-free image editing},
  author={Avrahami, Omri and Patashnik, Or and Fried, Ohad and Nemchinov, Egor and Aberman, Kfir and Lischinski, Dani and Cohen-Or, Daniel},
  booktitle={Proceedings of the Computer Vision and Pattern Recognition Conference},
  pages={7877--7888},
  year={2025}
}

@article{Geneval,
  title={Geneval: An object-focused framework for evaluating text-to-image alignment},
  author={Ghosh, Dhruba and Hajishirzi, Hannaneh and Schmidt, Ludwig},
  journal={Advances in Neural Information Processing Systems},
  volume={36},
  pages={52132--52152},
  year={2023}
}

@article{dpgbench,
  title={Ella: Equip diffusion models with llm for enhanced semantic alignment},
  author={Hu, Xiwei and Wang, Rui and Fang, Yixiao and Fu, Bin and Cheng, Pei and Yu, Gang},
  journal={arXiv preprint arXiv:2403.05135},
  year={2024}
}

@inproceedings{Coco,
  title={Microsoft coco: Common objects in context},
  author={Lin, Tsung-Yi and Maire, Michael and Belongie, Serge and Hays, James and Perona, Pietro and Ramanan, Deva and Doll{\'a}r, Piotr and Zitnick, C Lawrence},
  booktitle={European conference on computer vision},
  pages={740--755},
  year={2014},
  organization={Springer}
}

@inproceedings{celeb-a,
  title = {Deep Learning Face Attributes in the Wild},
  author = {Liu, Ziwei and Luo, Ping and Wang, Xiaogang and Tang, Xiaoou},
  booktitle = {Proceedings of International Conference on Computer Vision (ICCV)},
  month = {December},
  year = {2015} 
}

@article{celeba-hq,
  title={Progressive growing of gans for improved quality, stability, and variation},
  author={Karras, Tero and Aila, Timo and Laine, Samuli and Lehtinen, Jaakko},
  journal={arXiv preprint arXiv:1710.10196},
  year={2017}
}

@article{DAv2,
  title={Depth anything v2},
  author={Yang, Lihe and Kang, Bingyi and Huang, Zilong and Zhao, Zhen and Xu, Xiaogang and Feng, Jiashi and Zhao, Hengshuang},
  journal={Advances in Neural Information Processing Systems},
  volume={37},
  pages={21875--21911},
  year={2024}
}

@inproceedings{Dreambooth,
  title={Dreambooth: Fine tuning text-to-image diffusion models for subject-driven generation},
  author={Ruiz, Nataniel and Li, Yuanzhen and Jampani, Varun and Pritch, Yael and Rubinstein, Michael and Aberman, Kfir},
  booktitle={Proceedings of the IEEE/CVF conference on computer vision and pattern recognition},
  pages={22500--22510},
  year={2023}
}

@inproceedings{clipscore,
  title={Clipscore: A reference-free evaluation metric for image captioning},
  author={Hessel, Jack and Holtzman, Ari and Forbes, Maxwell and Le Bras, Ronan and Choi, Yejin},
  booktitle={Proceedings of the 2021 conference on empirical methods in natural language processing},
  pages={7514--7528},
  year={2021}
}

@inproceedings{clip,
  title={Learning transferable visual models from natural language supervision},
  author={Radford, Alec and Kim, Jong Wook and Hallacy, Chris and Ramesh, Aditya and Goh, Gabriel and Agarwal, Sandhini and Sastry, Girish and Askell, Amanda and Mishkin, Pamela and Clark, Jack and others},
  booktitle={International conference on machine learning},
  pages={8748--8763},
  year={2021},
  organization={PmLR}
}

@article{hpsv2,
  title={Human preference score v2: A solid benchmark for evaluating human preferences of text-to-image synthesis},
  author={Wu, Xiaoshi and Hao, Yiming and Sun, Keqiang and Chen, Yixiong and Zhu, Feng and Zhao, Rui and Li, Hongsheng},
  journal={arXiv preprint arXiv:2306.09341},
  year={2023}
}

@article{pickscore,
  title={Pick-a-pic: An open dataset of user preferences for text-to-image generation},
  author={Kirstain, Yuval and Polyak, Adam and Singer, Uriel and Matiana, Shahbuland and Penna, Joe and Levy, Omer},
  journal={Advances in neural information processing systems},
  volume={36},
  pages={36652--36663},
  year={2023}
}

@inproceedings{dino,
  title={Emerging properties in self-supervised vision transformers},
  author={Caron, Mathilde and Touvron, Hugo and Misra, Ishan and J{\'e}gou, Herv{\'e} and Mairal, Julien and Bojanowski, Piotr and Joulin, Armand},
  booktitle={Proceedings of the IEEE/CVF international conference on computer vision},
  pages={9650--9660},
  year={2021}
}

@article{dinov2,
  title={Dinov2: Learning robust visual features without supervision},
  author={Oquab, Maxime and Darcet, Timoth{\'e}e and Moutakanni, Th{\'e}o and Vo, Huy and Szafraniec, Marc and Khalidov, Vasil and Fernandez, Pierre and Haziza, Daniel and Massa, Francisco and El-Nouby, Alaaeldin and others},
  journal={arXiv preprint arXiv:2304.07193},
  year={2023}
}

@article{vendiscore,
  title={The vendi score: A diversity evaluation metric for machine learning},
  author={Friedman, Dan and Dieng, Adji Bousso},
  journal={arXiv preprint arXiv:2210.02410},
  year={2022}
}

@article{dreamsim,
  title={Dreamsim: Learning new dimensions of human visual similarity using synthetic data},
  author={Fu, Stephanie and Tamir, Netanel and Sundaram, Shobhita and Chai, Lucy and Zhang, Richard and Dekel, Tali and Isola, Phillip},
  journal={arXiv preprint arXiv:2306.09344},
  year={2023}
}

@inproceedings{sscd,
  title={A self-supervised descriptor for image copy detection},
  author={Pizzi, Ed and Roy, Sreya Dutta and Ravindra, Sugosh Nagavara and Goyal, Priya and Douze, Matthijs},
  booktitle={Proceedings of the IEEE/CVF Conference on Computer Vision and Pattern Recognition},
  pages={14532--14542},
  year={2022}
}

@article{FID,
  title={Gans trained by a two time-scale update rule converge to a local nash equilibrium},
  author={Heusel, Martin and Ramsauer, Hubert and Unterthiner, Thomas and Nessler, Bernhard and Hochreiter, Sepp},
  journal={Advances in neural information processing systems},
  volume={30},
  year={2017}
}

@inproceedings{clean-fid,
  title={On aliased resizing and surprising subtleties in gan evaluation},
  author={Parmar, Gaurav and Zhang, Richard and Zhu, Jun-Yan},
  booktitle={Proceedings of the IEEE/CVF conference on computer vision and pattern recognition},
  pages={11410--11420},
  year={2022}
}

@article{elhage2021mathematical,
   title={A Mathematical Framework for Transformer Circuits},
   author={Elhage, Nelson and Nanda, Neel and Olsson, Catherine and Henighan, Tom and Joseph, Nicholas and Mann, Ben and Askell, Amanda and Bai, Yuntao and Chen, Anna and Conerly, Tom and DasSarma, Nova and Drain, Dawn and Ganguli, Deep and Hatfield-Dodds, Zac and Hernandez, Danny and Jones, Andy and Kernion, Jackson and Lovitt, Liane and Ndousse, Kamal and Amodei, Dario and Brown, Tom and Clark, Jack and Kaplan, Jared and McCandlish, Sam and Olah, Chris},
   year={2021},
   journal={Transformer Circuits Thread},
   note={https://transformer-circuits.pub/2021/framework/index.html}
}



\clearpage
\appendix

\section*{\texorpdfstring{\centering Supplementary Materials}{Supplementary Materials}}
\section*{Table of Contents}
\begin{enumerate}[label=\Alph*.]
    \item \hyperlink{sec:addn-results}{\textbf{Additional Results}} \dotfill \pageref{sec:addn-results}
    \item \hyperlink{sec:addn-models}{\textbf{Results on QwenImage, Sana1.5}} \dotfill \pageref{sec:addn-models}
    \item \hyperlink{sec:klein}{\textbf{Results on Step Distilled Model (Flux-Klein)}} \dotfill \pageref{sec:klein}
    \item \hyperlink{sec:implementation}{\textbf{Implementation Details}} \dotfill \pageref{sec:implementation}
    \item \hyperlink{sec:addn-baselines}{\textbf{Additional Baselines}} \dotfill \pageref{sec:addn-baselines}
    \item \hyperlink{sec:addn-dataset}{\textbf{Results on DPG-Bench}} \dotfill \pageref{sec:addn-dataset}
    \item \hyperlink{sec:tradeoff}{\textbf{Diversity vs Prompt Adherence Tradeoff}} \dotfill \pageref{sec:tradeoff}
    \item \hyperlink{sec:ablation}{\textbf{Ablation Studies}} \dotfill \pageref{sec:ablation}
\end{enumerate}

\setcounter{section}{0}
\setcounter{figure}{0}
\setcounter{table}{0}
\renewcommand{\thesection}{\Alph{section}}
\renewcommand{\thefigure}{\Alph{figure}}
\renewcommand{\thetable}{\Alph{table}}

\section{Additional Results}
\label{sec:addn-results}
\hypertarget{sec:addn-results}{}

We present additional qualitative results for text-to-image generation with FLUX in Fig.~\ref{appendix-fig:addn-t2i-flux-outputs1}, ~\ref{appendix-fig:addn-t2i-flux-outputs2} \&~\ref{appendix-fig:addn-t2i-flux-outputs3}, unconditional generation with UNet-based DDPM in Fig.~\ref{appendix-fig:addn-ddpm} and image conditioned generation in Fig.~\ref{appendix-fig:addn-i2i-depth-outputs} \&~\ref{appendix-fig:addn-i2i-kontext-outputs}.

\section{Results on QwenImage, Sana1.5}
\label{sec:addn-models}
\hypertarget{sec:addn-models}{}

\begin{figure}[H]
  \centering
   \includegraphics[width=0.8\linewidth]{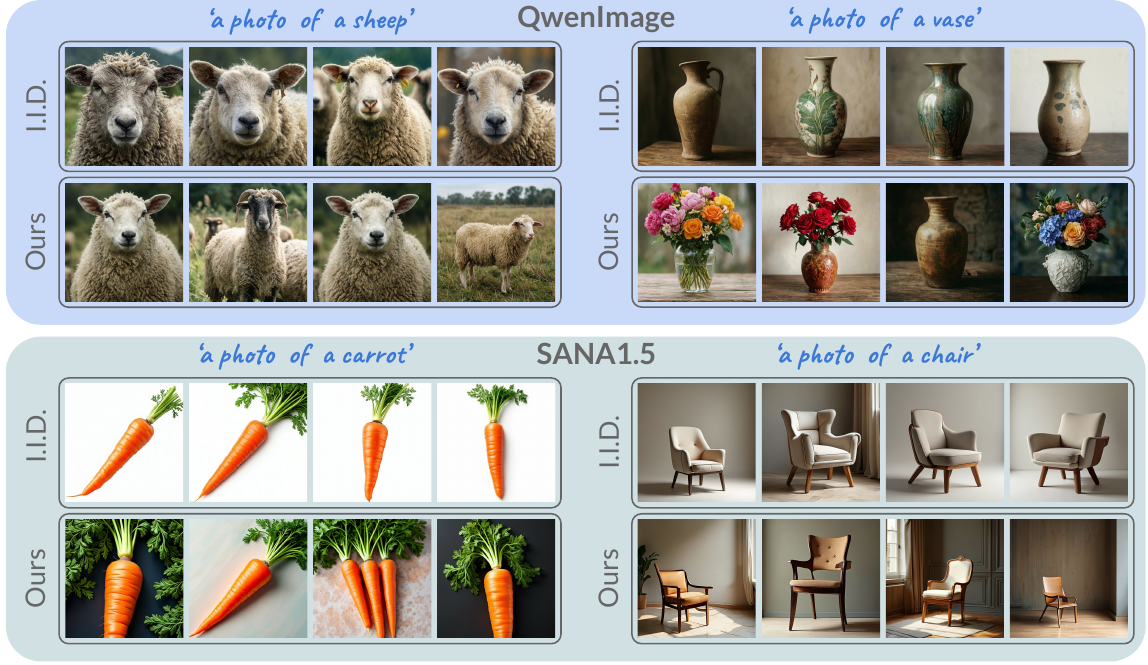}
   \vspace{-2mm}
   \caption{\textbf{Qualitative results for QwenImage and Sana1.5.}}
   \label{appendix-fig:additional-t2i}
\end{figure}

\begin{figure}[!p]
  \centering
   \includegraphics[width=0.95\linewidth]{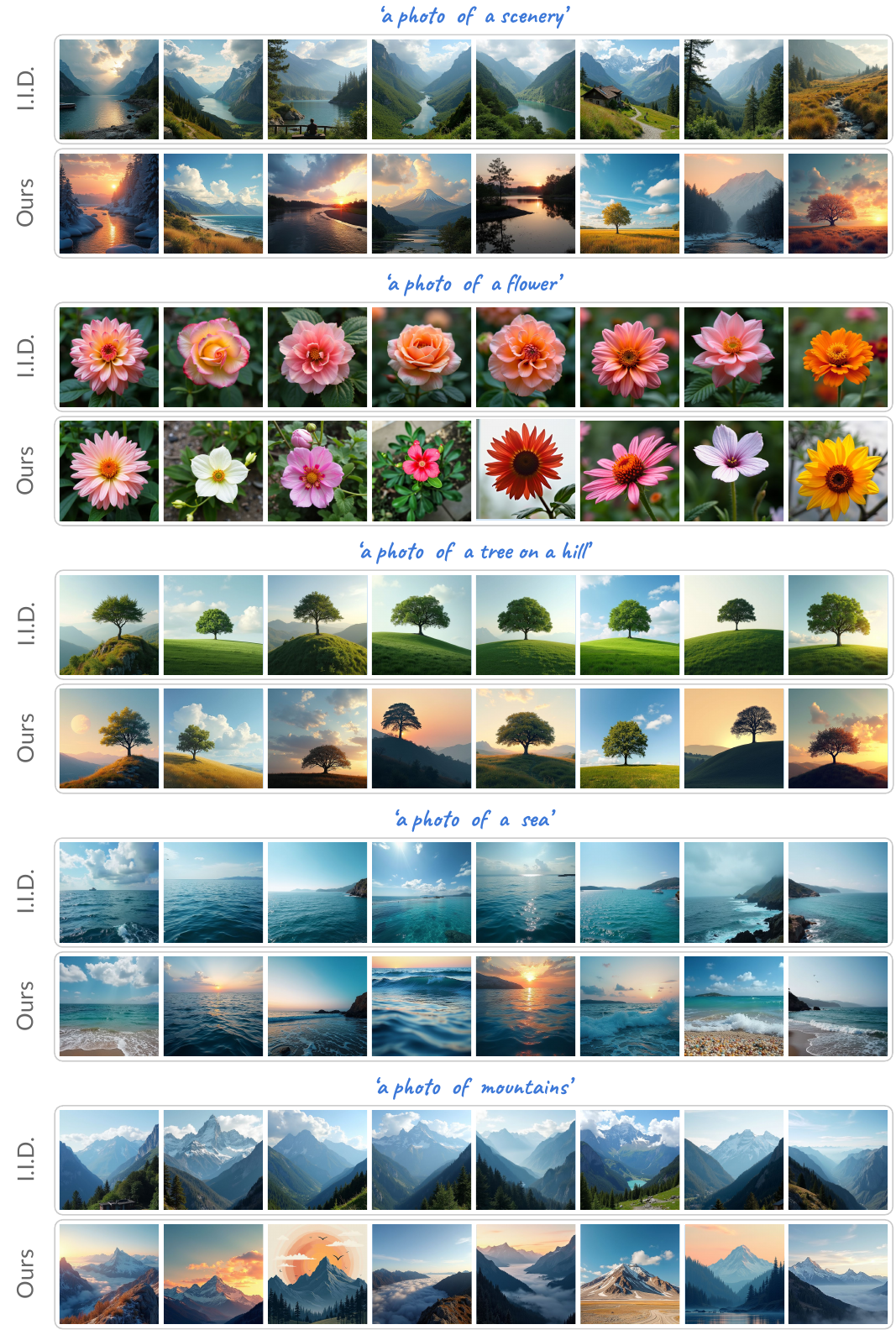}
   \caption{\textbf{Qualitative results for text-to-image generation with FLUX.1-dev.} Our method yields remarkably greater diversity in layouts and colors when handling prompts centered on natural elements\textemdash{}such as landscapes, vegetation, and water bodies when compared to I.I.D. sampling.}
   \label{appendix-fig:addn-t2i-flux-outputs1}
\end{figure}

\begin{figure}[!p]
  \centering
   \includegraphics[width=0.95\linewidth]{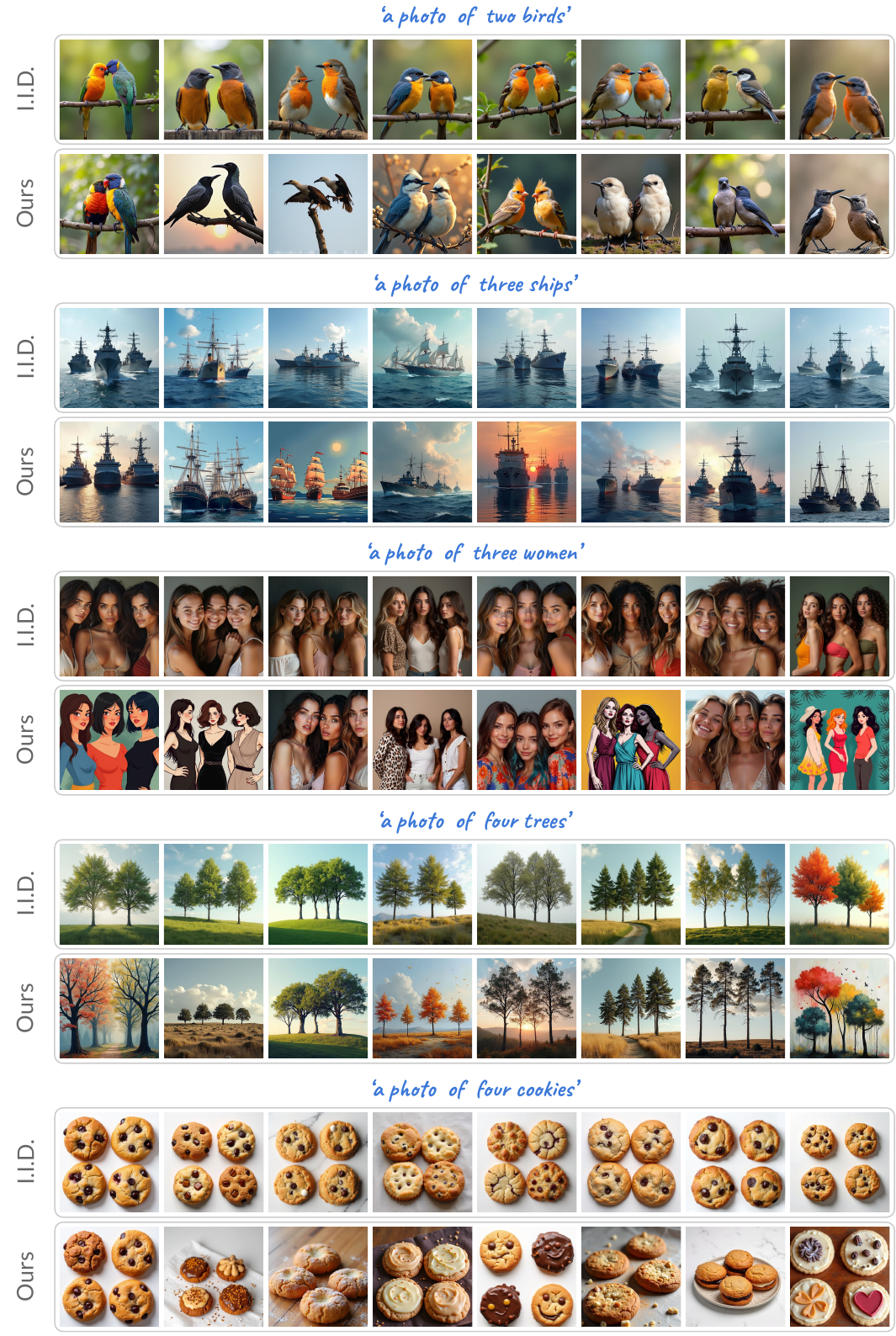}
   \caption{\textbf{Qualitative results for text-to-image generation with FLUX.1-dev.} Our method demonstrates greater compositional diversity on prompts involving object numeracy, without sacrificing accuracy in generating the requested number of elements.}
   \label{appendix-fig:addn-t2i-flux-outputs2}
\end{figure}

\begin{figure}[!p]
  \centering
   \includegraphics[width=0.95\linewidth]{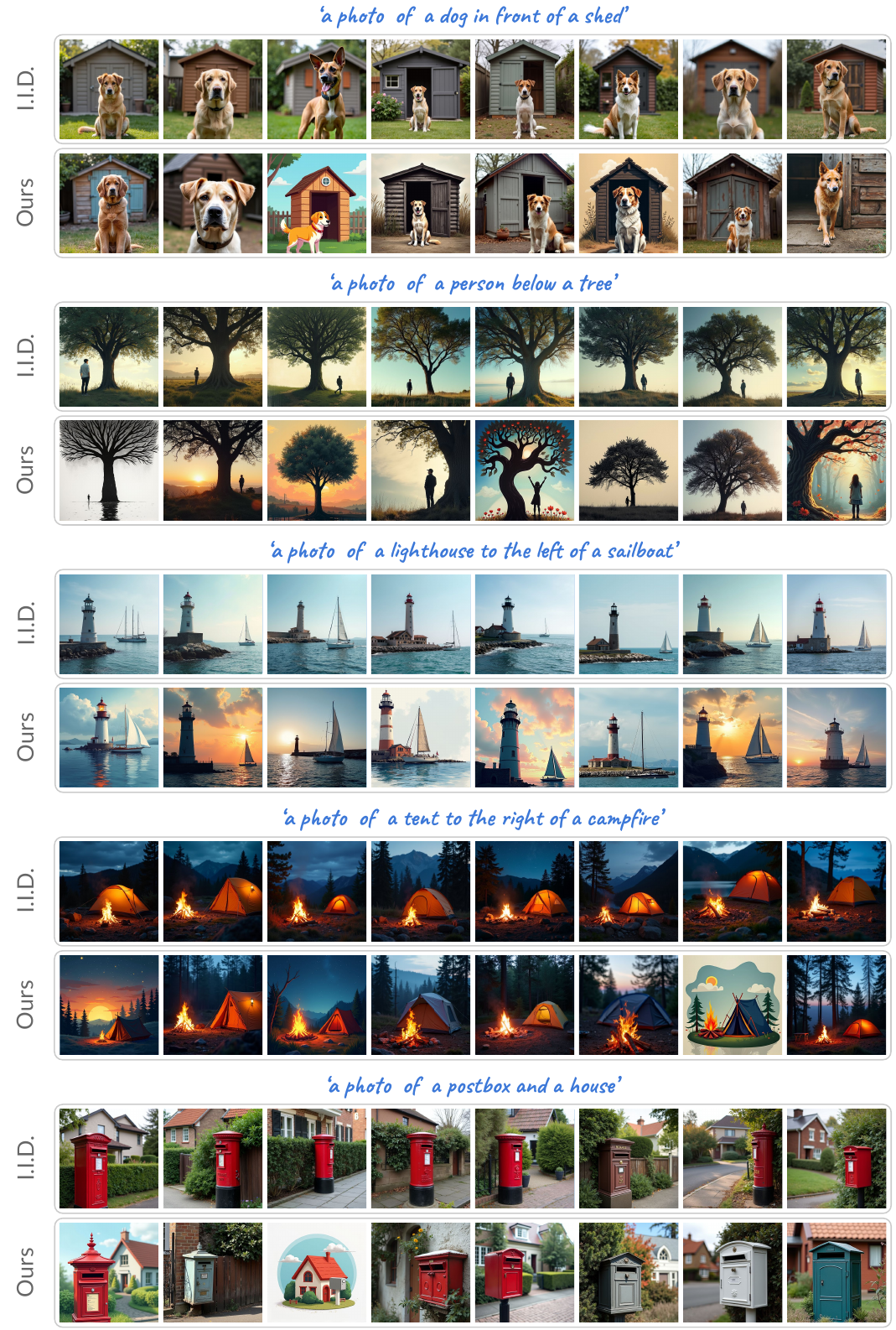}
   \caption{\textbf{Qualitative results for text-to-image generation with FLUX.1-dev.} Even when handling complex spatial prompts, our approach significantly boosts generation diversity while maintaining precise spatial layout.}
   \label{appendix-fig:addn-t2i-flux-outputs3}
\end{figure}

\begin{figure}[t]
  \centering
   \includegraphics[width=1.0\linewidth]{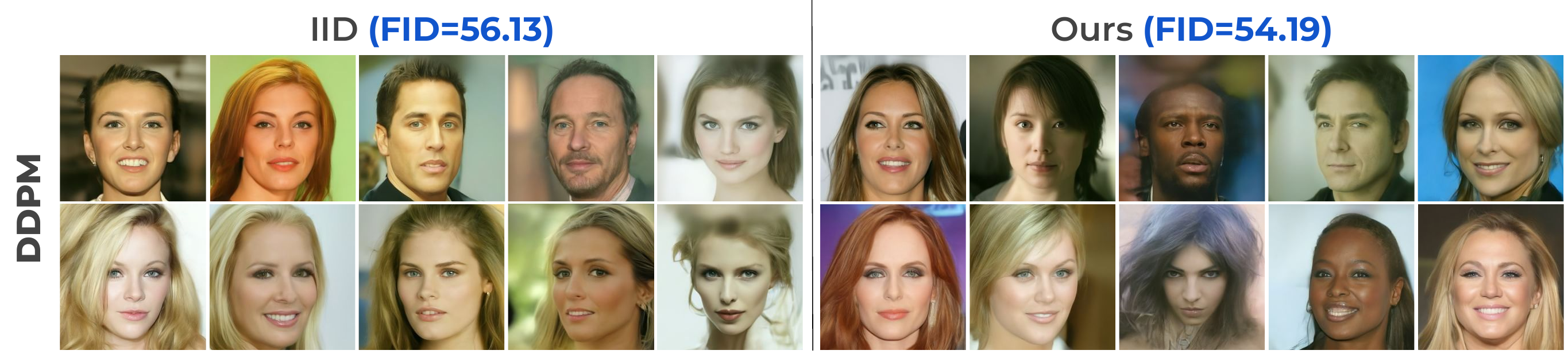}
   \vspace{-6mm}
   \caption{\textbf{Human face generation results using the UNet-based DDPM model.} Our method successfully generalizes beyond DiT-based architectures to deliver higher sample diversity and this is validated quantitatively by a lower FID score.}
   \vspace{-4mm}
   \label{appendix-fig:addn-ddpm}
\end{figure}

\begin{figure}[H]
  \centering
   \includegraphics[width=0.95\linewidth]{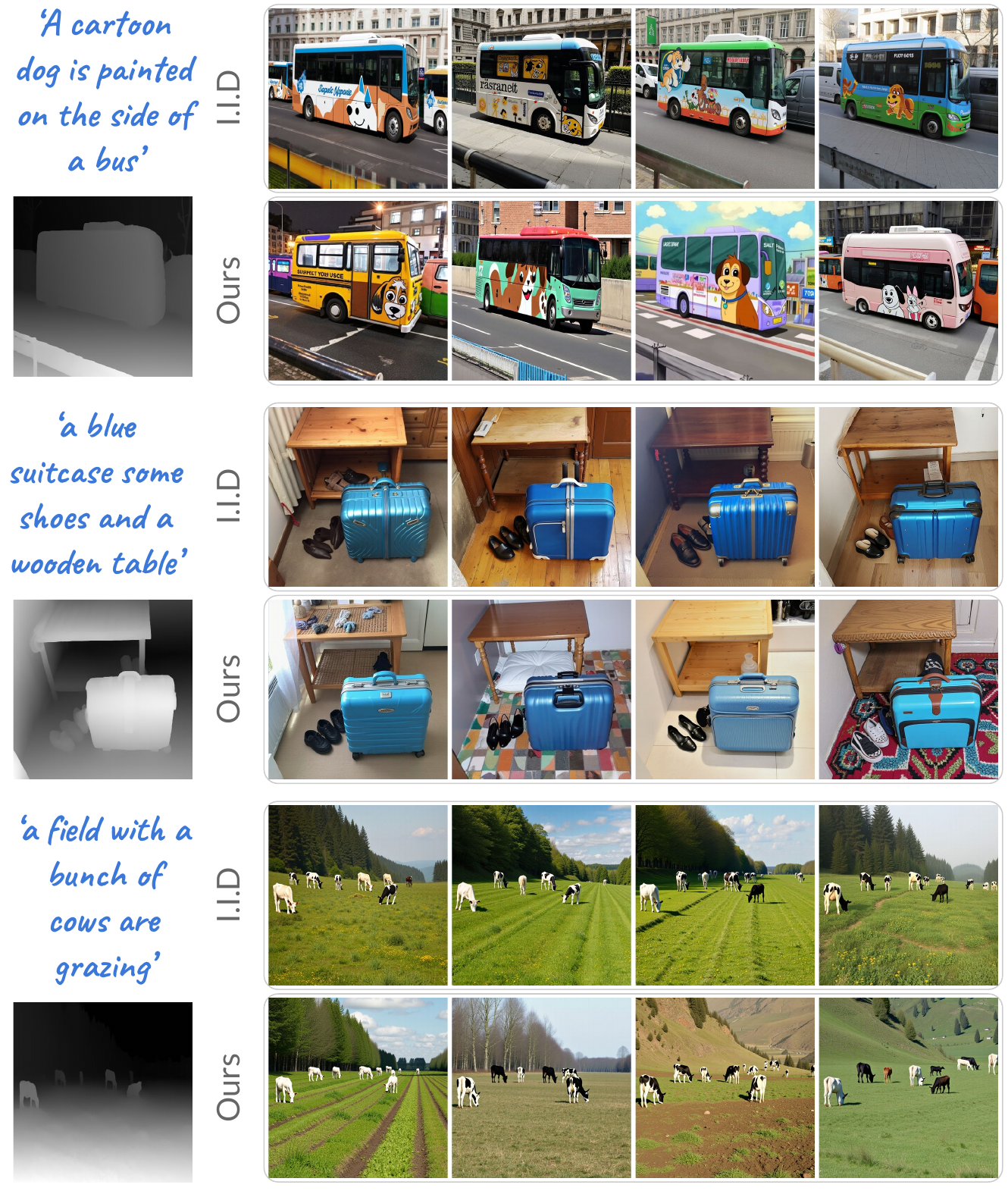}
   \vspace{-2mm}
   \caption{\textbf{Qualitative results for depth-to-image generation using FLUX-Depth.} Our method generates diverse backgrounds and textures while strictly adhering to strong spatial conditioning\textemdash{}such as depth maps.}
   \label{appendix-fig:addn-i2i-depth-outputs}
\end{figure}

\begin{figure}[H]
  \centering
   \includegraphics[width=0.95\linewidth]{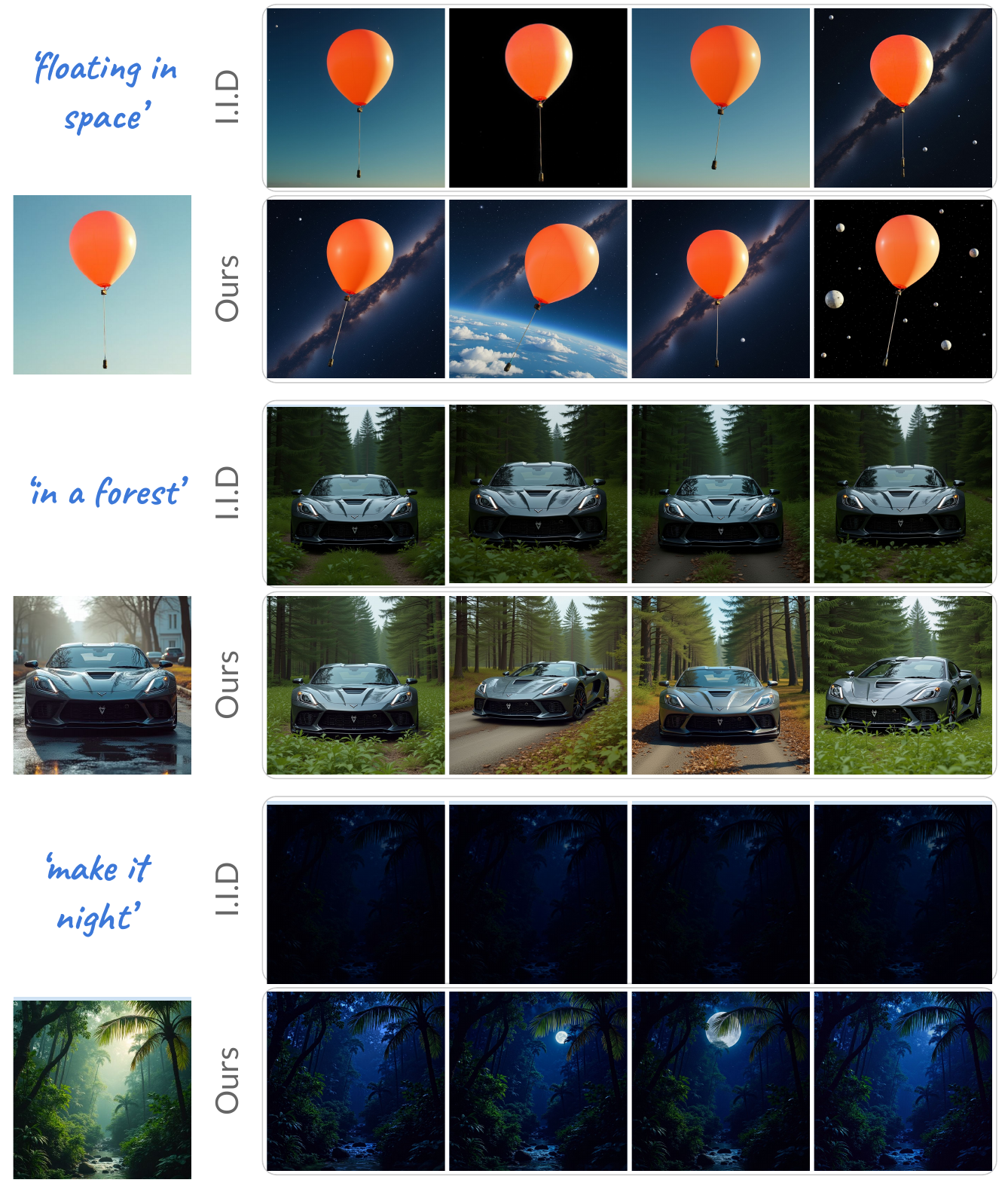}
   \caption{\textbf{Qualitative results for image-to-image generation with FLUX.1-Kontext.} By faithfully preserving reference concepts and unlocking diverse element compositions, our method gives users an expanded pool of unique image choices.}
   \label{appendix-fig:addn-i2i-kontext-outputs}
   \vspace{-2mm}
\end{figure}

We demonstrate generalization of our approach by implementing our method on additional flow-based T2I generation models - QwenImage~\cite{qwen-image} and Sana-1.5~\cite{sana} in Tab.~\ref{tab:additional-t2i-models} and Fig.~\ref{appendix-fig:additional-t2i}. Our method improves the diversity of both models, with a marginal change in the text alignment. 

\vspace{1mm}
\begin{table}[h]
\centering
\caption{Results on Geneval with additional T2I models - QwenImage and Sana1.5}
\vspace{-2mm}
\resizebox{\textwidth}{!}{
\begin{tabular}{c|ccccc|ccc}
\hline
Model & Method & DINO$\uparrow$ & VS$\uparrow$ & DreamSim$\uparrow$ & MSS$\downarrow$ & CLIPScore$\uparrow$ & HPSv2$\uparrow$ & PickScore$\uparrow$ \\ \hline
  & IID    &   0.65\tiny{$\pm$0.09}     &  3.50\tiny{$\pm$1.35}    & 0.37\tiny{$\pm$0.12}   &   0.19\tiny{$\pm$0.08}  &   31.33\tiny{$\pm$4.14}  &  \textbf{0.28\tiny{$\pm$0.04}}      &    \textbf{22.57\tiny{$\pm$1.41}} \\
\multirow{-2}{*}{Qwen-Image} &
  \cellcolor[HTML]{90EE90}Ours &
  \cellcolor[HTML]{90EE90}\textbf{0.67\tiny{$\pm$0.09}} &
  \cellcolor[HTML]{90EE90}\textbf{3.65\tiny{$\pm$1.41}} &
  \cellcolor[HTML]{90EE90}\textbf{0.38\tiny{$\pm$0.13}} &
  \cellcolor[HTML]{90EE90}\textbf{0.18\tiny{$\pm$0.07}} &
  \cellcolor[HTML]{90EE90}\textbf{31.38\tiny{$\pm$3.94}} &
  \cellcolor[HTML]{90EE90}0.27\tiny{$\pm$0.04} &
  \cellcolor[HTML]{90EE90}22.40\tiny{$\pm$1.34} \\ \hline
      & IID & 0.45\tiny{$\pm$0.10} & 1.93\tiny{$\pm$0.61} & 0.17\tiny{$\pm$0.07} & 0.46\tiny{$\pm$0.10} & \textbf{33.27\tiny{$\pm$3.46}} & 0.31\tiny{$\pm$0.03} & \textbf{23.28\tiny{$\pm$1.07}} \\
\multirow{-2}{*}{Sana1.5} &
  \cellcolor[HTML]{90EE90}Ours &
  \cellcolor[HTML]{90EE90}\textbf{0.50\tiny{$\pm$0.10}} &
  \cellcolor[HTML]{90EE90}\textbf{2.14\tiny{$\pm$0.73}} &
  \cellcolor[HTML]{90EE90}\textbf{0.20\tiny{$\pm$0.08}} &
  \cellcolor[HTML]{90EE90}\textbf{0.41\tiny{$\pm$0.10}} &
  \cellcolor[HTML]{90EE90}33.09\tiny{$\pm$3.59} &
  \cellcolor[HTML]{90EE90}\textbf{0.31\tiny{$\pm$0.03}} &
  \cellcolor[HTML]{90EE90}22.97\tiny{$\pm$1.16} \\ \hline
\end{tabular}
}
\label{tab:additional-t2i-models}
\end{table}

\vspace{-2mm}

\begin{figure}[!p]
  \centering
   \includegraphics[width=0.95\linewidth]{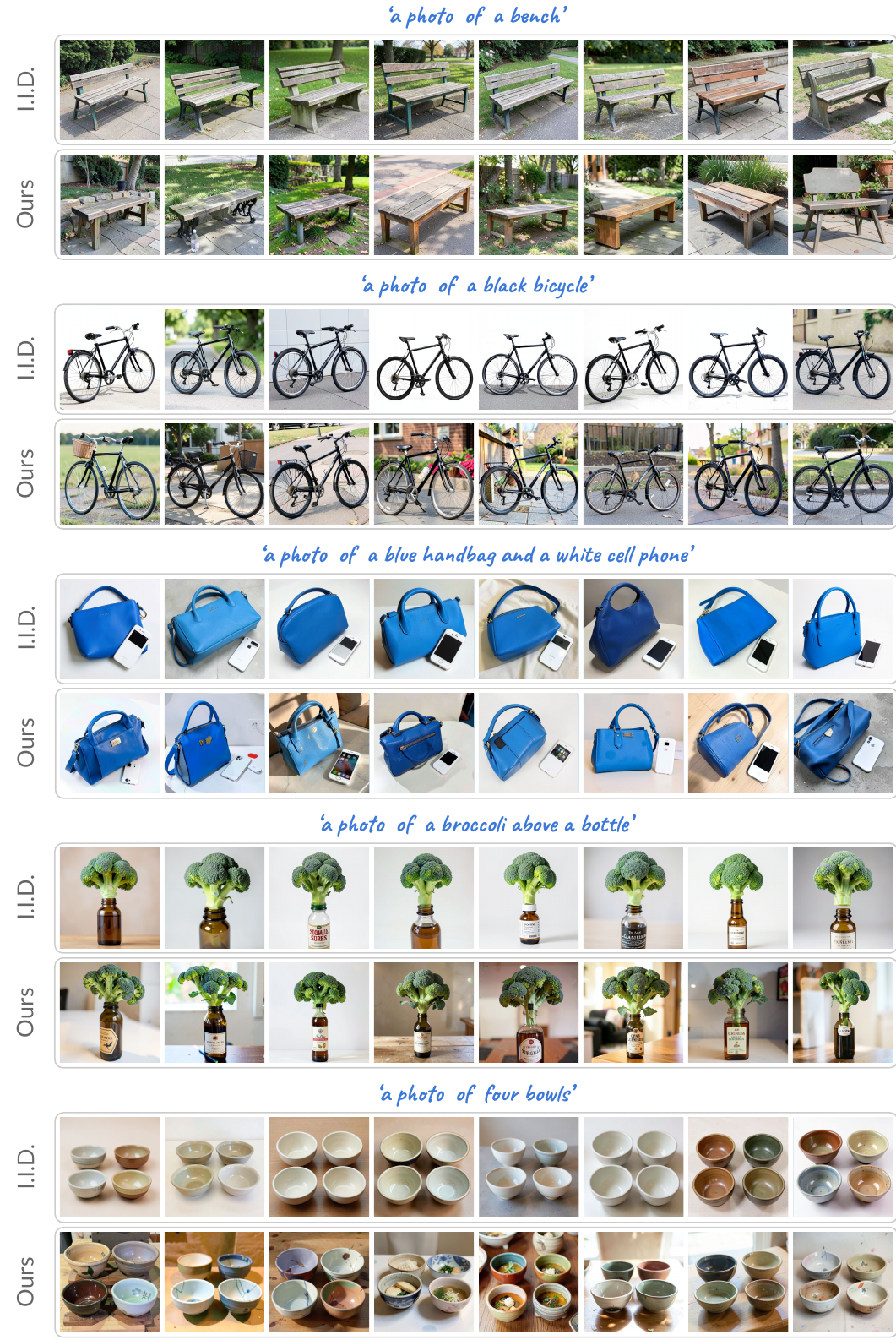}
   \caption{\textbf{Results for $4$ step text-to-image generation with FLUX.2-Klein-4B.} Our method enhances sample diversity even in step-distilled models, demonstrating the robustness and generalization of our feature-space intervention.}
   \label{appendix-fig:klein_t2i}
\end{figure}

\section{Results on step distilled flow model - FLUX.2-Klein-4B} 
\label{sec:klein}
\hypertarget{sec:klein}{}

Our method can be seamlessly integrated with step-distilled flow models. To demonstrate this generalization, we implement our method on FLUX.2-Klein-4B~\cite{flux-2} for text-to-image generation with $4$ steps. We present results in Fig.~\ref{appendix-fig:klein_t2i} and  Tab.~\ref{tab:geneval-klein-t2i}. Notably, our method generates the most diverse results while preserving text-alignment and image quality.

\vspace{2mm}
\begin{table}[h]
\centering
\caption{Results of T2I diverse samplers on Geneval with FLUX.2-Klein-4B.}
\vspace{-2mm}
\resizebox{\textwidth}{!}{%
\begin{tabular}{c | c | cccc | ccc }
\hline
Method & Latency$\downarrow$ & DINO$\uparrow$ & VS$\uparrow$ & DreamSim$\uparrow$ & MSS$\downarrow$ & CLIPScore$\uparrow$ & HPSv2$\uparrow$ & PickScore$\uparrow$  \\ \hline

IID & \textbf{0.14s} & 0.46\tiny{$\pm$0.10}  &  2.14\tiny{$\pm$0.79} &  0.19\tiny{$\pm$0.08} & 0.41\tiny{$\pm$0.12} & \textbf{33.78\tiny{$\pm$3.62}} & \textbf{0.31\tiny{$\pm$0.03}} & \textbf{23.50\tiny{$\pm$1.093}} \\

\rowcolor[HTML]{D6D6D6} 
Particle Guidance~\cite{particle-guidance} & 0.45s & 0.48\tiny{$\pm$0.10}  & 2.36\tiny{$\pm$0.87} & 0.22\tiny{$\pm$0.08} & 0.36\tiny{$\pm$0.10} & 33.71\tiny{$\pm$3.61} &  0.30\tiny{$\pm$0.03} & 23.37\tiny{$\pm$1.11}  \\

CADS~\cite{cads} & 0.14s & 0.50\tiny{$\pm$0.02} & 2.42\tiny{$\pm$0.20} & 0.09\tiny{$\pm$0.01} & 0.51\tiny{$\pm$0.02} & 19.01\tiny{$\pm$2.75} & 0.13\tiny{$\pm$0.02} & 17.27\tiny{$\pm$0.84} \\

\rowcolor[HTML]{D6D6D6}
Shielded Diffusion~\cite{shielded-diff} & 0.15s & 0.55\tiny{$\pm$0.08} & 2.45\tiny{$\pm$0.90} & 0.23\tiny{$\pm$0.08} & 0.33\tiny{$\pm$0.08} &  33.77\tiny{$\pm$3.55} & 0.30\tiny{$\pm$0.03} &  23.19\tiny{$\pm$1.14}\\

CNO~\cite{CNO} & 0.23s & 0.55\tiny{$\pm$0.09} & 2.42\tiny{$\pm$0.92} & 0.24\tiny{$\pm$0.09} & 0.32\tiny{$\pm$0.10} &  33.79\tiny{$\pm$3.63} & 0.30\tiny{$\pm$0.03} & 23.43\tiny{$\pm$1.10}  \\

\rowcolor[HTML]{D6D6D6}
GroupInference~\cite{group-inference} (8 of 64) & 1.14s & 0.48\tiny{$\pm$0.10} & 2.35\tiny{$\pm$0.87} & 0.18\tiny{$\pm$0.08} & 0.46\tiny{$\pm$0.12} &  33.54\tiny{$\pm$3.51} & 0.28\tiny{$\pm$0.04} & 22.82\tiny{$\pm$1.20} \\

\rowcolor[HTML]{D6D6D6}
GroupInference~\cite{group-inference} (8 of 128) & 1.64s & 0.51\tiny{$\pm$0.10} & 2.52\tiny{$\pm$0.98}  & 0.20\tiny{$\pm$0.09} & 0.44\tiny{$\pm$0.12} &  33.51\tiny{$\pm$3.52} & 0.28\tiny{$\pm$0.04} & 22.74\tiny{$\pm$1.22} \\ \hline

\rowcolor[HTML]{90EE90} 
Ours & 0.16s & \textbf{0.60\tiny{$\pm$0.07}} & \textbf{2.73\tiny{$\pm$0.97}} & \textbf{0.26\tiny{$\pm$0.08}} & \textbf{0.27\tiny{$\pm$0.07}} & 33.27\tiny{$\pm$3.67} &  0.30\tiny{$\pm$0.03} &  23.09\tiny{$\pm$1.14} \\ \hline
\end{tabular}
}
\label{tab:geneval-klein-t2i}
\end{table}

\section{Implementation Details}
\label{sec:implementation}
\hypertarget{sec:implementation}{}
In this section, we provide implementation details about the baselines, datasets, and evaluation metrics. 

\subsection{Baselines}
\label{subsec:baselines}

\noindent{\textbf{I.I.D.}} For Text-to-Image (T2I) Generation using FLUX.1-dev~\cite{flux}, we use 28 inference timesteps with a guidance scale of 3.5. For FLUX.2-Klein-4B~\cite{flux-2}, we use 4 inference timesteps and a guidance scale of 0. For Depth-to-Image (D2I) Generation using FLUX.1-Depth-dev~\cite{flux}, we use 28 inference timesteps with a guidance scale of 10, and for Image-to-Image (I2I) Generation, we use 28 inference timesteps and a guidance scale of 3.5. All generations are performed at a resolution of $512^2$ with an initial seed of 42. We use the same inference settings for all baselines unless explicitly mentioned otherwise. For all quantitative comparisons, we generate 8 images per prompt. For the UNet-based DDPM model~\cite{ddpm}, we use 50 inference timesteps, an initial seed of 42, and generate 5000 images at a resolution of $256^2$ to compute FID~\cite{FID}.

\vspace{2mm}
\noindent{\textbf{Interval Guidance}}~\cite{interval-guidance} We apply guidance during the interval $t \in [0.9,0.1]$. This baseline is not applicable to FLUX.2-Klein-4B as it does not use guidance.

\vspace{2mm}
\noindent{\textbf{Particle Guidance}}~\cite{particle-guidance} For Particle Guidance, we consider a Coefficient of 150 using the feature space potential obtained with DINO-VIT-s/8~\cite{dino} feature extractor.

\vspace{2mm}
\noindent{\textbf{Shielded Diffusion}}~\cite{shielded-diff} We follow the guidelines recommended by the authors to find an appropriate repellence radius by calculating the median pairwise $L_2$ distances of the generated latents at the final timestep. We then tested 16 values from 0 to 2 times this distance, resulting in a repellence radius, $r = 160$. We also use the overcompensation factor provided by the authors, $\lambda = 1.6$.

\vspace{2mm}
\noindent{\textbf{CNO}}~\cite{CNO} Following the authors, We use $\gamma=1.0$, window size,$w=16$, learning rate,$\eta=0.01$. We use 12 optimization steps for FLUX.1-dev and 3 optimization steps for FLUX.2-Klein-4B.

\vspace{2mm}
\noindent{\textbf{Group Inference}}~\cite{group-inference} We consider 128 candidate samples as well as 64 candidate samples and filter samples using reward functions at a pruning ratio $\rho = 0.5$ for non-timestep distilled models and $\rho=0.9$ for the timestep distilled model, FLUX.2-Klein-4B. We use a weighting factor, $\lambda = 1.5$ in both cases and use CLIPScore~\cite{clipscore} as the quality reward function along with DINOScore~\cite{dinov2} as the diversity reward function. We sample images using 20 inference steps as recommended to minimize latency.

\vspace{2mm}
\noindent{\textbf{Ours}} We use a consistent temperature of $\tau = 0.5$ across all generations. The optimization hyperparameters for T2I generation are varied based on the sample batch size, $N$ (as detailed in Tab.~\ref{tab:hyperparams}) and the specific model choice of the model (as shown in Tab.~\ref{tab:model_hyperparams}). Empirically, we find that the best results are obtained by scaling up the learning rate, $w$ and decreasing interpolation coefficient, $\beta$ as the batch size increases. Our algorithm is applied during the initial timesteps $t \in [1.0,0.8]$, targeting the second MMDiT block for FLUX.1 models and the first block for FLUX.2-Klein-4B. 

\noindent Leveraging insights from prior work on inference-time intervention in UNet-based models~\cite{h-space,balancing-act}, we apply our method to the middle bottleneck layer of the UNet, commonly referred to as the "\textit{h-space}". While our setup is broadly similar to the $N=8$ configuration for FLUX.1-dev, we specifically adjust the learning rate, $w$ to $0.02$ and the interpolation coefficient, $\beta$ to $0.2$.

\vspace{2mm}
\begin{table}[h]
\centering
\caption{\textbf{Hyperparameter list for various batch sizes,N for FLUX.1-dev}}
\vspace{-2mm}
\begin{tabular}{c|c|c|c|c}
\textbf{Hyperparameter} & \textbf{N=8} & \textbf{N=16} & \textbf{N=64} & \textbf{N=128} \\ \hline
Optimization steps, $N_{opt}$ & 30 & 25 & 20 & 30 \\
Learning Rate, $w$ & 0.5 & 0.75 & 2.25 & 2\\
Temperature, $\tau$ & 0.5 & 0.5 & 0.5 & 0.5 \\
Interpolation Coefficient, $\beta$ & 0.4 & 0.2 & 0.15 & 0.05 \\
Block, $B_i$ & 2 & 2 & 2 & 2 \\
Start Timestep, $T_a$ & $1.0$ & $1.0$ & $1.0$ & $1.0$ \\
Stop Timestep, $T_b$ & $0.8$ & $0.8$ & $0.8$ & 0.8 \\ \hline               
\end{tabular}
\label{tab:hyperparams}
\end{table}

\vspace{2mm}
\begin{table}[h]
\centering
\caption{\textbf{Hyperparameter list for various FLUX models for Batch Size,N=8}}
\vspace{-2mm}
\resizebox{\textwidth}{!}{
\begin{tabular}{c|c|c|c|c}
\textbf{Hyperparameter} & \textbf{FLUX.1-dev} & \textbf{FLUX.1-Depth-dev} & \textbf{FLUX.1-Kontext-dev} & \textbf{FLUX.2-Klein-4B} \\ \hline
Optimization Steps, $N_{opt}$ & 30 & 20 & 20 & 10 \\
Learning rate, $w$ & 0.5 & 0.5 & 2 & 0.05 \\
Temperature, $\tau$ & 0.5 & 0.5 & 0.5 & 0.5 \\
Interpolation Coefficient, $\beta$ & 0.4 & 0.1 & 0.1 & 0.1 \\
Block, $B_i$ & 2 & 2 & 2 & 1 \\
Start Timestep, $T_a$  & 1.0 & 1.0 & 1.0 & 1.0 \\
Stop Timestep, $T_b$  & 0.8 & 0.8 & 0.8 & 0.8 \\ \hline
\end{tabular}
}
\label{tab:model_hyperparams}
\end{table}

\subsection{Datasets}
\label{subsec:datasets}

\noindent\textbf{Text-to-image (T2I) Generation.} We use the Geneval~\cite{Geneval} dataset, an object-focused evaluation benchmark, to evaluate the diversity of T2I generations using FLUX.1-dev and FLUX.2-Klein-4B. We use all 553 prompts and sample 8 images per prompt. Additionally, we report results on the DPGBench~\cite{dpgbench}, which consists of dense prompts in order to show the robustness of our method.

\vspace{2mm}
\noindent\textbf{Depth-to-image (D2I) Generation.} We report results on D2I generations using 500 examples from the validation split of COCO 2017~\cite{Coco} and generate 8 samples per image. More specifically, we obtain depth maps for the images using DepthAnythingv2-Large~\cite{DAv2} and generate depth-conditioned images using FLUX.1-Depth-dev.

\vspace{2mm}
\noindent\textbf{Image-to-image (I2I) Generation.} To benchmark personalized image generation, we use the Dreambooth~\cite{Dreambooth} dataset. We use one sample image per category for all 30 categories and use all 25 prompts, giving us 750 image-prompt pairs. We generate 8 samples per pair using FLUX.1-Kontext-dev.  

\vspace{2mm}
\noindent\textbf{Human Face (Unconditional) Generation.} To benchmark unconditional image generation using UNet-based DDPM, we follow~\cite{ddpm} and use the Celeb-A-HQ~\cite{celeb-a,celeba-hq} dataset as the reference distribution. We consider all 30$\textit{K}$ images of the reference set and generate 5$\textit{K}$ images for comparison.

\subsection{Metrics}
\label{subsec:metrics}

\textbf{Diversity Metrics.} To evaluate the visual diversity and sample variance, we employ the following metrics:

\vspace{2mm}
\noindent\textbf{DINOScore}~\cite{dinov2} We use the DINOv2~\cite{dinov2} encoder to extract patch-level features of the generated samples and then compute the average pairwise cosine distance between all features. Lower values indicate similar images and values close to 1 indicate high diversity.

\vspace{2mm}
\noindent\textbf{VendiScore}~\cite{vendiscore} We compute a similarity kernel using normalized [CLS] token embeddings obtained using DINOv2~\cite{dinov2}. Following~\cite{vendiscore}, we report the exponential of the Shannon entropy of the kernel's eigenvalues. This metric ranges from [1,N], where N is the number of samples per prompt; a score of 1 indicates identity, while a score of 8 (in our setup) signifies complete uniqueness.

\vspace{2mm}
\noindent\textbf{DreamSim}~\cite{dreamsim} To capture perceptual mid-level similarities, we compute pairwise DreamSim cosine dissimilarity using a DINO-ViT-B/16 backbone fine-tuned on human perceptual judgments~\cite{dreamsim}. Values near 0 indicate perceptual similarity, while higher values reflect increased diversity.

\vspace{2mm}
\noindent\textbf{MSS}~\cite{cads} Following~\cite{cads}, we assess sample similarity using features from a self-supervised copy detection model (SSCD)~\cite{sscd}. Values close to 0 indicate high diversity, whereas values closer to 1 indicate low diversity.

\vspace{2mm}
\noindent\textbf{FID}~\cite{FID} We assess sample diversity of human faces using clean-fid~\cite{clean-fid}. Lower FID indicates a closer match to the reference distribution i.e. higher diversity.

\vspace{4mm}
\noindent\textbf{Alignment and Quality Metrics.} To evaluate text alignment and sample quality, we utilize the following metrics:

\vspace{2mm}
\noindent\textbf{CLIPScore}~\cite{clipscore}. We measure text alignment by calculating the cosine similarity between the text prompt and image embeddings of the samples using a pretrained CLIP ViT-B/32~\cite{clip} model. Lower values indicate poor text alignment and vice-versa.

\vspace{2mm}
\noindent\textbf{HPSv2}~\cite{hpsv2}. Beyond semantic adherence, we evaluate human preference alignment using this metric. It uses a CLIP ViT-H/14~\cite{clip} backbone fine-tuned to predict human choices.

\vspace{2mm}
\noindent\textbf{PickScore}~\cite{pickscore}. We evaluate how likely the generated samples would be "picked" by the users using this metric. Similar to HPSv2, it makes use of a CLIP ViT-H/14 encoder fine-tuned on a dataset of human choices.

\section{Additional Baselines}
\label{sec:addn-baselines}
\hypertarget{sec:addn-baselines}{}

We compare with two additional baselines that focus on inference time improvement in the diversity of the Flow models in Tab.~\ref{tab:geneval-t2i}. 

\vspace{2mm} 
\noindent\textbf{Perturbing the internal MMDiT features.} This is a simple baseline (also visualized in Fig.2 of the main paper), where we add Gaussian perturbations to the internal MMDiT features. Though this leads to improvement in diversity over i.i.d, it degrades the text alignment. Further, improving the strength of the perturbation will result in even further degradation of the text alignment and artifacts in generated images, as shown in Fig.2 of the main paper. 

\vspace{2mm}
\noindent\textbf{CADS}~\cite{cads}. We implement the existing CADS baseline where the text embedding is perturbed at each step of denoising. As suggested, more noise is added in the earlier timestep to mitigate collapse, and no noise is added for subsequent timesteps for stronger text-alignment. We perturb the text embeddings with a Gaussian noise scale, $s$ of $0.25$ during the denoising process with $\tau_2 =0.9$ and $\tau_1 = 0.6$ and a mixing factor, $\psi=1.0$.

\vspace{2mm}
\begin{table}[h]
\centering
\caption{Comparison of T2I diverse samplers with FLUX.1-dev on Geneval~\cite{Geneval}.}
\vspace{-2mm}
\resizebox{\textwidth}{!}{%
\begin{tabular}{c | c | cccc | ccc }
\hline
Method & Latency$\downarrow$ & DINO$\uparrow$ & VS$\uparrow$ & DreamSim$\uparrow$ & MSS$\downarrow$ & CLIPScore$\uparrow$ & HPSv2$\uparrow$ & PickScore$\uparrow$  \\ \hline

IID & \textbf{1.59s} & 0.57\tiny{$\pm$0.10}  &  2.77\tiny{$\pm$1.06} &  0.27\tiny{$\pm$0.10} & 0.32\tiny{$\pm$0.10} & 32.43\tiny{$\pm$3.71} & 0.30\tiny{$\pm$0.03} & \textbf{23.37\tiny{$\pm$1.14}} \\

\rowcolor[HTML]{D6D6D6} 
Particle Guidance~\cite{particle-guidance} & 2.94s & 0.58\tiny{$\pm$0.11}  & 3.16\tiny{$\pm$1.17} & 0.31\tiny{$\pm0.12$} & 0.29\tiny{$\pm$0.11} &  32.22\tiny{$\pm$3.90} & 0.30\tiny{$\pm$0.04} & 23.13\tiny{$\pm$1.31}  \\

Interval Guidance~\cite{interval-guidance} & 1.59s & 0.57\tiny{$\pm$0.10} & 2.88\tiny{$\pm$1.11} & 0.28\tiny{$\pm$0.10} & 0.300\tiny{$\pm$0.10} & 32.46\tiny{$\pm$3.75} & 0.30\tiny{$\pm$0.03} & 23.27\tiny{$\pm$1.14} \\

\rowcolor[HTML]{D6D6D6}
CADS~\cite{cads} & 1.59s & \textbf{0.85\tiny{$\pm$0.01}} & \textbf{7.31\tiny{$\pm$0.37}} & \textbf{0.69\tiny{$\pm$0.04}} & \textbf{0.09\tiny{$\pm$0.03}} & 18.05\tiny{$\pm$3.05} & 0.16\tiny{$\pm$0.04} &  18.34\tiny{$\pm$1.11}\\

Shielded Diffusion~\cite{shielded-diff} & 1.59s & 0.61\tiny{$\pm$0.10} & 3.35\tiny{$\pm$1.22} & 0.40\tiny{$\pm$0.13} & 0.23\tiny{$\pm$0.08} & 32.11\tiny{$\pm$3.89} &  0.28\tiny{$\pm$0.04} &  22.80\tiny{$\pm$1.35}\\

\rowcolor[HTML]{D6D6D6}
CNO~\cite{CNO} & 1.75s & 0.63\tiny{$\pm$0.08} & 3.46\tiny{$\pm$1.10} & 0.39\tiny{$\pm$0.10} & 0.22\tiny{$\pm$0.07} & 31.97\tiny{$\pm$3.77} & 0.28\tiny{$\pm$0.04} & 22.67\tiny{$\pm$1.34} \\

GroupInference~\cite{group-inference} (8 of 64) & 2.86s & 0.67\tiny{$\pm$0.07} & 3.37\tiny{$\pm$1.17} & 0.35\tiny{$\pm$0.10} & 0.21\tiny{$\pm$0.07} & \textbf{32.49\tiny{$\pm$3.74}} & 0.31\tiny{$\pm$0.03} &  23.34\tiny{$\pm$1.20}\\

GroupInference~\cite{group-inference} (8 of 128) & 4.55s & 0.70\tiny{$\pm$0.07} & 3.61\tiny{$\pm$1.26} & 0.37\tiny{$\pm$0.11} & 0.19\tiny{$\pm$0.07} & 32.37\tiny{$\pm$3.77} & \textbf{0.31\tiny{$\pm$0.03}} &  23.28\tiny{$\pm$1.21}\\

\rowcolor[HTML]{D6D6D6}
Feature Perturbation & 1.59s & 0.64\tiny{$\pm$0.08} & 3.56\tiny{$\pm$1.20} & 0.33\tiny{$\pm$0.10} & 0.25\tiny{$\pm$0.07} & 31.74\tiny{$\pm$3.92} & 0.29\tiny{$\pm$0.04} & 22.41\tiny{$\pm$1.39} \\  \hline

\rowcolor[HTML]{90EE90} 
Ours & 1.70s & 0.68\tiny{$\pm$0.07} & 3.57\tiny{$\pm$1.17} & 0.40\tiny{$\pm$0.09} & 0.19\tiny{$\pm$0.06} & 32.05\tiny{$\pm$3.64} &  0.29\tiny{$\pm$0.03} &  22.92\tiny{$\pm$1.20} \\ \hline
\end{tabular}
}
\label{tab:geneval-t2i}
\end{table} 

\begin{figure}[h]
  \centering
   \includegraphics[width=0.95\linewidth]{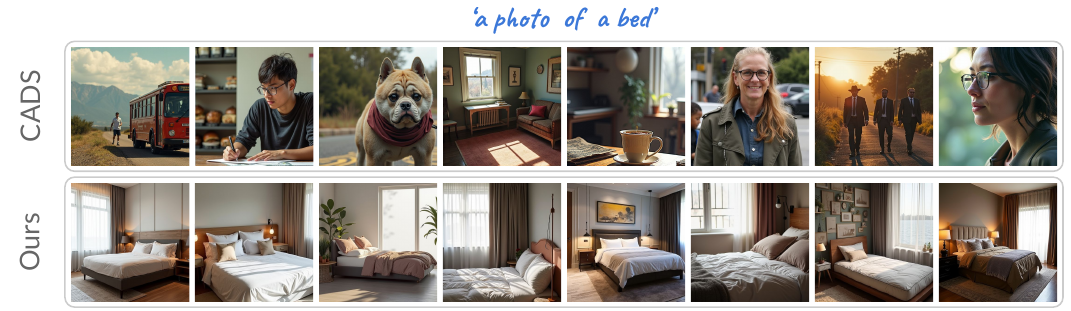}
   \caption{\textbf{Comparing with CADS.} CADS achieves diversity by ignoring user prompts, whereas our method produces high diversity without sacrificing prompt alignment.}
   \label{appendix-fig:cads}
\end{figure}  

\section{Results on additional DPG-Bench}
\label{sec:addn-dataset}
\hypertarget{sec:addn-dataset}{}

We present results on an additional benchmark - DPGBench~\cite{dpgbench} in Tab.~\ref{tab:dpg-t2i}. Our method consistently improves over all the baselines except CADS in diversity. CADS has a substantially low CLIPScore ($17.62$ vs $33.62$), indicating it sacrifices text alignment for improving the diversity. 

\vspace{1mm}
\begin{table}[h]
\centering
\caption{Comparison on DPGBench of T2I diverse samplers with FLUX.1-dev.}
\vspace{-2mm}
\resizebox{\textwidth}{!}{%
\begin{tabular}{c | c | cccc | ccc }
\hline
Method & Latency$\downarrow$ & DINO$\uparrow$ & VS$\uparrow$ & DreamSim$\uparrow$ & MSS$\downarrow$ &  CLIPScore$\uparrow$ & HPSv2$\uparrow$ & PickScore$\uparrow$  \\ \hline

IID & \textbf{1.59s} & 0.53\tiny{$\pm$0.08}  &  2.35\tiny{$\pm$0.73} &  0.20\tiny{$\pm$0.07} & 0.41\tiny{$\pm$0.09} & \textbf{33.65\tiny{$\pm$3.63}} & \textbf{0.29\tiny{$\pm$0.04}} & \textbf{22.34\tiny{$\pm$1.20}} \\

\rowcolor[HTML]{D6D6D6} 
Particle Guidance~\cite{particle-guidance} & 2.94s & 0.58\tiny{$\pm$0.08}  & 2.54\tiny{$\pm$0.03} & 0.25\tiny{$\pm$0.09} & 0.36\tiny{$\pm$0.10} &  33.45\tiny{$\pm$3.77} & 0.28\tiny{$\pm$0.04} & 22.14\tiny{$\pm$1.20}  \\

Interval Guidance~\cite{interval-guidance} & 1.59s & 0.53\tiny{$\pm$0.08} & 2.49\tiny{$\pm$0.78} & 0.21\tiny{$\pm$0.07} & 0.39\tiny{$\pm$0.09} & 33.59\tiny{$\pm$3.60} & 0.28\tiny{$\pm$0.04} & 22.18\tiny{$\pm$1.21} \\

\rowcolor[HTML]{D6D6D6} 
CADS~\cite{cads} & 1.59s & \textbf{0.83\tiny{$\pm$0.03}} & \textbf{7.05\tiny{$\pm$0.6}} & \textbf{0.63\tiny{$\pm$0.06}} & \textbf{0.11\tiny{$\pm$0.03}} & 17.62\tiny{$\pm$4.86} & 0.16\tiny{$\pm$0.04} &  18.30\tiny{$\pm$1.41}\\

Shielded Diffusion~\cite{shielded-diff} & 1.59s & 0.58\tiny{$\pm$0.08} & 2.54\tiny{$\pm$0.83} & 0.25\tiny{$\pm$0.09} & 0.36\tiny{$\pm$0.10} & 33.45\tiny{$\pm$3.77} &  0.28\tiny{$\pm$0.04} &  22.14\tiny{$\pm$1.28}\\

\rowcolor[HTML]{D6D6D6} 
CNO~\cite{CNO} & 1.75s & 0.58\tiny{$\pm$0.07} & 2.62\tiny{$\pm$0.81} & 0.24\tiny{$\pm$0.08} & 0.35\tiny{$\pm$0.09} & 33.48\tiny{$\pm$3.8} & 0.28\tiny{$\pm$0.04} & 22.00\tiny{$\pm$1.25} \\

GroupInference~\cite{group-inference} (8 of 64) & 2.86s & 0.55\tiny{$\pm$0.08} & 2.56\tiny{$\pm$0.78} & 0.23\tiny{$\pm$0.07} & 0.36\tiny{$\pm$0.08} & 33.77\tiny{$\pm$3.55} & 0.29\tiny{$\pm$0.04} &  22.18\tiny{$\pm$1.22}\\

GroupInference~\cite{group-inference} (8 of 128) & 4.55s & {0.62\tiny{$\pm$0.07}} & {2.65\tiny{$\pm$0.82}} & 0.25\tiny{$\pm$0.08} & 0.34\tiny{$\pm$0.08} & 33.29\tiny{$\pm$3.62} & 0.28\tiny{$\pm$0.04} &  22.13\tiny{$\pm$1.22}\\ \hline

\rowcolor[HTML]{90EE90} 
Ours & 1.70s & 0.64\tiny{$\pm$0.06} & 2.87\tiny{$\pm$0.86} & {0.28\tiny{$\pm$0.08}} & {0.30\tiny{$\pm$0.08}} & 33.62\tiny{$\pm$3.61} &  0.28\tiny{$\pm$0.04} &  22.07\tiny{$\pm$1.22} \\ \hline
\end{tabular}
}
\label{tab:dpg-t2i}
\end{table} 

\section{Diversity vs Prompt Adherence Tradeoff}
\label{sec:tradeoff}
\hypertarget{sec:tradeoff}{}

\begin{figure}[h]
  \centering
   \includegraphics[width=1\linewidth]{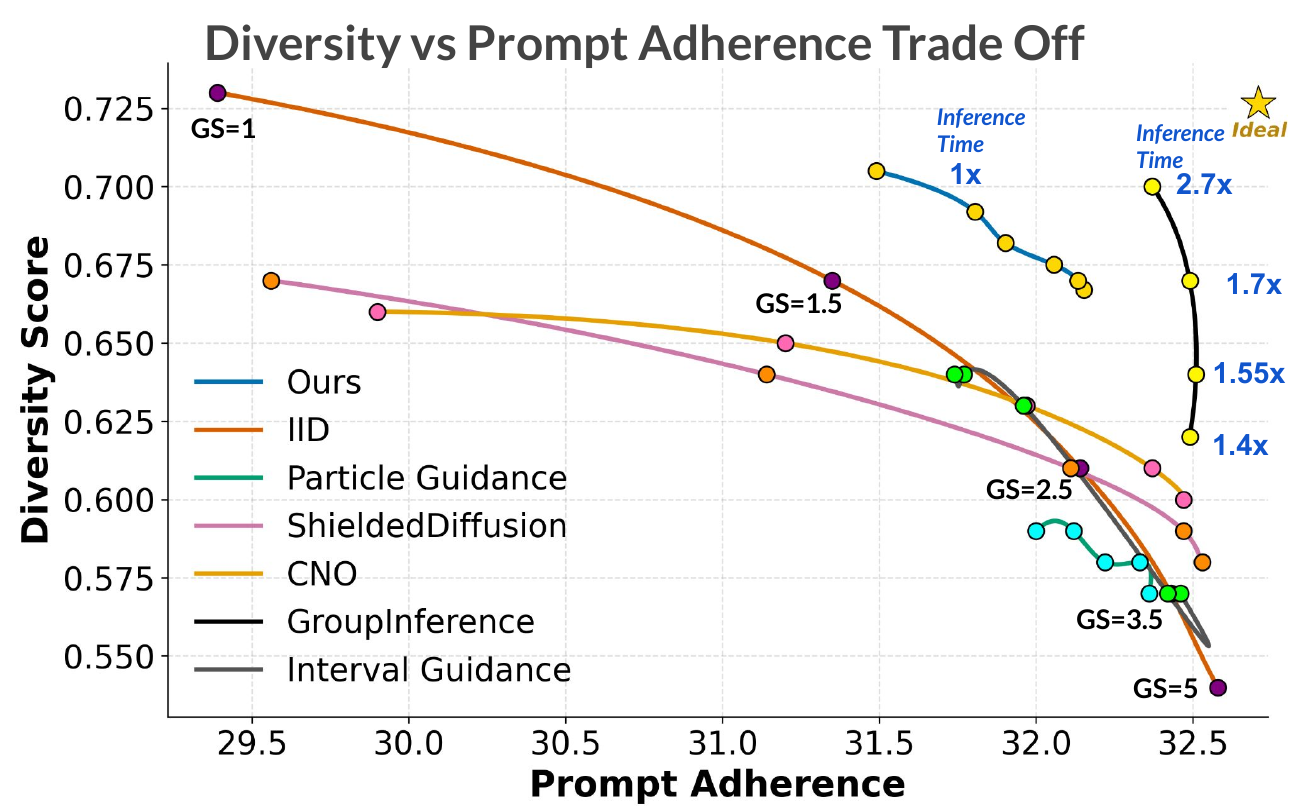}
    \vspace{-4mm} 
   \caption{\textbf{Diversity vs Prompt Adherence Tradeoff}: By selecting different hyperparameter choices for the baselines, we plot the diversity vs prompt adherence tradeoff. Our method shows much higher diversity with good prompt adherence compared to most of the baselines. Group Inference is several times slower than our method.}
    \vspace{-2mm}
   \label{appendix-fig:tradeoff}
\end{figure}

We present tradeoff between diversity and prompt adherence for baselines in Fig.~\ref{appendix-fig:tradeoff} by varying their hyperparameters. Notably, Group Inference is several times during inference than our method. 

\section{Ablations}
\label{sec:ablation}
\hypertarget{sec:ablation}{} 

\begin{figure}[h]
  \centering
   \includegraphics[width=0.75\linewidth]{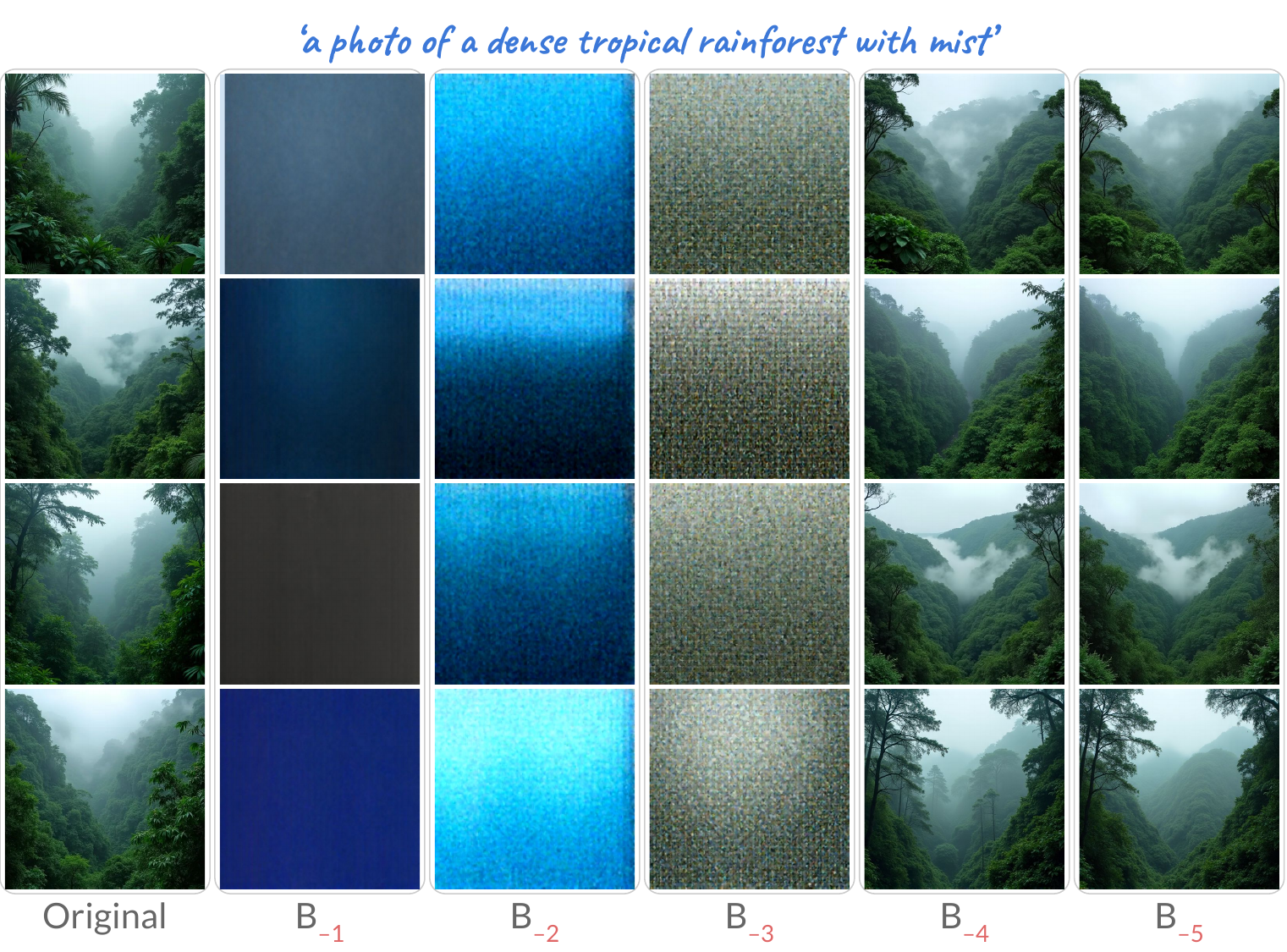}
   \caption{\textbf{Effect of Removing MMDiT Blocks $B_i$ during inference}: Bypassing blocks $B_1,B_2,B_3$ results in generation of noise, revealing their importance for generation. Bypassing further blocks such as $B_4,B_5$ only causes minor change in the outputs.}
   \label{appendix-fig:1_kontext}
\end{figure}

We ablate over the important design choices for hyperparameters, \textbf{a)} Choice of Block and \textbf{b)} Choice of Timesteps for our Feature Self-Guidance algorithm below:

\vspace{2mm}
\noindent\textbf{Choice of MMDiT Block.}
We performed an insightful experiment, where we skipped a particular MMDiT block during inference of the FLUX.1 model in Fig.~\ref{appendix-fig:1_kontext}. Removing the initial blocks (1-3) completely corrupts the generation, whereas removing the later blocks (4-5) does not have a large effect. Hence, we apply our guidance in the earlier block of the model. Further, to select the best block, we perform an ablation over blocks for applying the dispersive loss in Tab.~\ref{tab:block-ablation}. Applying guidance on block 2 features is most effective in improving the diversity and retaining the text-to-image alignment of the samples.  

\vspace{2mm}
\begin{table}[h]
\centering
\caption{\textbf{T2I Block Ablation.} While earlier blocks offer greater diversity at the expense of text-image alignment, later blocks exhibit the opposite trend. Applying our method to block $B_2$ yields the optimal trade-off between diversity and alignment.}
\resizebox{\textwidth}{!}{%
\begin{tabular}{c | cccc | ccc }
\hline
Block & DINO$\uparrow$ & VS$\uparrow$ & DreamSim$\uparrow$ & MSS$\downarrow$ & CLIPScore$\uparrow$ & HPSv2$\uparrow$ & PickScore$\uparrow$  \\ \hline

Block, $B_1$ &  \textbf{0.69\tiny{$\pm$0.07}}  &  \textbf{3.91\tiny{$\pm$1.18}} & \textbf{0.44\tiny{$\pm$0.10}} & \textbf{0.18\tiny{$\pm$0.06}} & 31.29\tiny{$\pm$3.64} & 0.28\tiny{$\pm$0.04} & 22.22\tiny{$\pm$1.40} \\

\rowcolor[HTML]{90EE90} 
Block, $B_2$ & 0.68\tiny{$\pm$0.07} & 3.57\tiny{$\pm$1.17} & 0.40\tiny{$\pm$0.09} & 0.19\tiny{$\pm$0.06} & 32.05\tiny{$\pm$3.64} &  0.29\tiny{$\pm$0.03} &  22.92\tiny{$\pm$1.20} \\

Block, $B_3$ & 0.60\tiny{$\pm$0.09} & 2.94\tiny{$\pm$1.11} & 0.29\tiny{$\pm$0.10} & 0.29\tiny{$\pm$0.09} & 32.39\tiny{$\pm$3.68} & 0.30\tiny{$\pm$0.03} & 23.37\tiny{$\pm$1.15} \\

\rowcolor[HTML]{D6D6D6}
Block, $B_4$ & 0.60\tiny{$\pm$0.09} & 2.88\tiny{$\pm$1.08} & 0.28\tiny{$\pm$0.10} & 0.29\tiny{$\pm$0.09} & 32.38\tiny{$\pm$3.67} & 0.30\tiny{$\pm$0.03} &  23.40\tiny{$\pm$1.15}\\

Block, $B_5$ & 0.59\tiny{$\pm$0.09} & 2.83\tiny{$\pm$1.07} & 0.28\tiny{$\pm$0.10} & 0.30\tiny{$\pm$0.09} & \textbf{32.37\tiny{$\pm$3.77}} &  \textbf{0.31\tiny{$\pm$0.03}} &  \textbf{23.42\tiny{$\pm$1.16}} \\ \hline
\end{tabular}
}
\label{tab:block-ablation}
\end{table} 

\vspace{2mm}

\begin{figure}[h]
  \centering
   \includegraphics[width=0.8\linewidth]{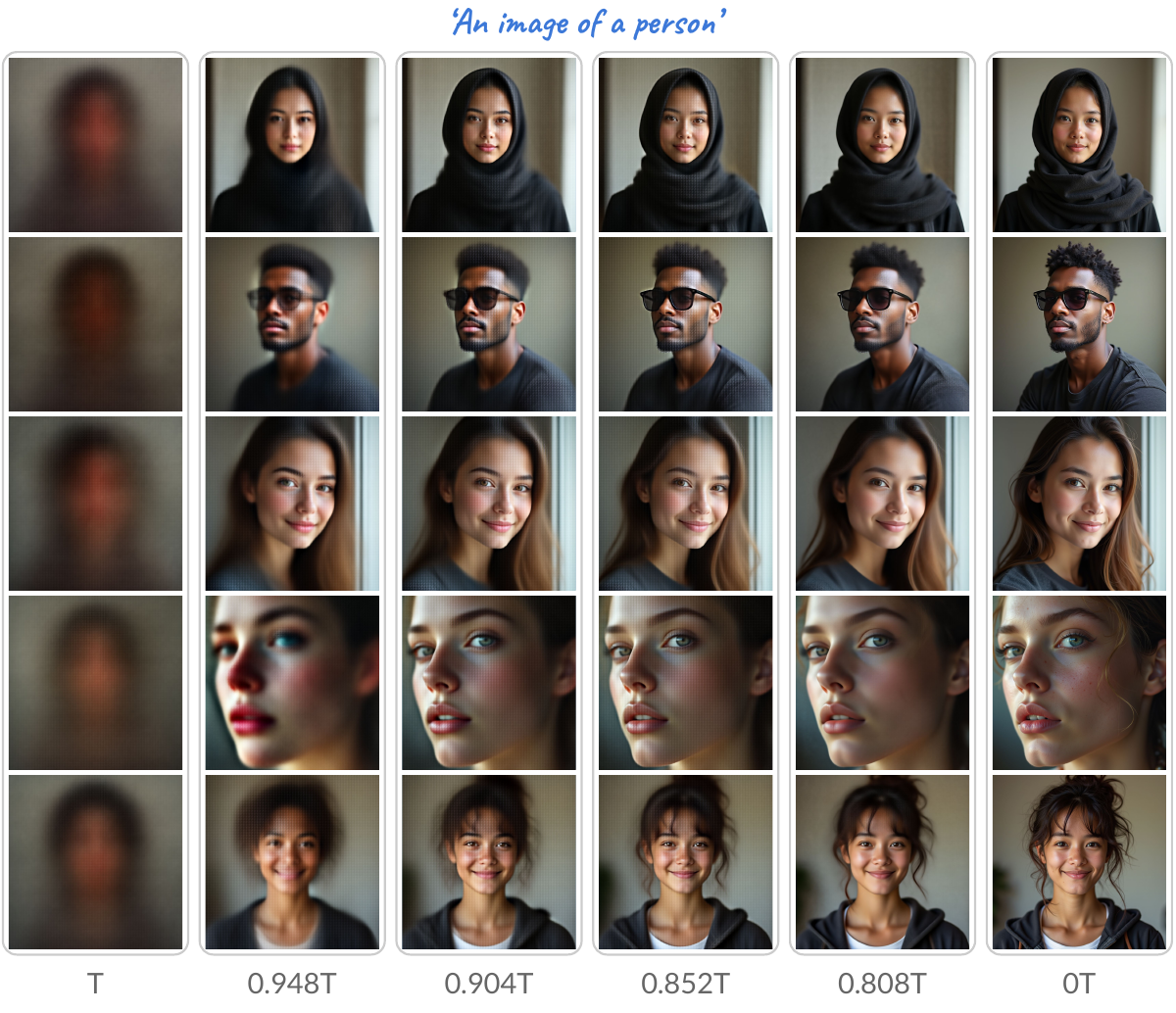}
   \vspace{-4mm}
   \caption{\textbf{$x_0$ predictions at intermediate timesteps}: Structure of the image is largely formed within $t \in [1.0,0.8]$, and finer details are added in the remaining timesteps.}
   \label{appendix-fig:1_t2i_timestep}
\end{figure}

\vspace{2mm}
\noindent\textbf{Choice of Timesteps Guidance.}
We ablate the choice of timestep for applying the proposed feature self-guidance in  Tab.~\ref{tab:timestep-ablation} and Fig.~\ref{appendix-fig:1_t2i_timestep}.  Applying the guidance from T=1.0 to T=0.8 provides good diversity and prompt alignment.

\vspace{2mm}
\begin{table}[H]
\centering
\caption{\textbf{T2I Timestep Ablation.} Early guidance termination improves text-image alignment, whereas extended guidance yields diminishing returns in sample diversity at a high cost to alignment. We find that T=1.0 to T=0.8 is the optimal duration.}
\vspace{-2mm}
\resizebox{\textwidth}{!}{%
\begin{tabular}{c | cccc | ccc }
\hline
Timestep & DINO$\uparrow$ & VS$\uparrow$ & DreamSim$\uparrow$ & MSS$\downarrow$ & CLIPScore$\uparrow$ & HPSv2$\uparrow$ & PickScore$\uparrow$  \\ \hline

Stop Timestep, $T_b=0.9$ & 0.66\tiny{$\pm$0.07} & 3.39\tiny{$\pm$1.18} & 0.36\tiny{$\pm$0.09} & 0.22\tiny{$\pm$0.07} & \textbf{32.23\tiny{$\pm$3.67}} &  \textbf{0.30\tiny{$\pm$0.03}} &  \textbf{23.12\tiny{$\pm$1.19}}\\

\rowcolor[HTML]{90EE90} 
Stop Timestep, $T_b=0.8$ & \textbf{0.68\tiny{$\pm$0.07}} & 3.57\tiny{$\pm$1.17} & 0.40\tiny{$\pm$0.09} & 0.19\tiny{$\pm$0.06} & 32.05\tiny{$\pm$3.64} &  0.29\tiny{$\pm$0.03} &  22.92\tiny{$\pm$1.20} \\

Stop Timestep, $T_b=0.7$ & 0.68\tiny{$\pm$0.07}  & 3.59\tiny{$\pm$1.16} & 0.40\tiny{$\pm$0.09} & 0.19\tiny{$\pm$0.06} &  31.97\tiny{$\pm$3.60} & 0.29\tiny{$\pm$0.03} & 22.81\tiny{$\pm$1.21}  \\

\rowcolor[HTML]{D6D6D6}
Stop Timestep, $T_b=0.5$ & 0.68\tiny{$\pm$0.07} & \textbf{3.62\tiny{$\pm$1.15}} & \textbf{0.41\tiny{$\pm$0.08}} & \textbf{0.19\tiny{$\pm$0.05}} & 31.82\tiny{$\pm$3.59} & 0.29\tiny{$\pm$0.03} &  22.66\tiny{$\pm$1.22}\\ \hline
\end{tabular}
}
\label{tab:timestep-ablation}
\end{table}  

%
%

\end{document}